\documentclass{egpubl}
\usepackage{eg2024}
 
\ConferenceSubmission 
\usepackage[T1]{fontenc}
\usepackage{dfadobe}  
\biberVersion
\BibtexOrBiblatex
\usepackage[backend=biber,bibstyle=EG,citestyle=alphabetic,backref=true,giveninits=true]{biblatex} 
\electronicVersion
\PrintedOrElectronic
\ifpdf \usepackage[pdftex]{graphicx} \pdfcompresslevel=9
\else \usepackage[dvips]{graphicx} \fi

\usepackage{egweblnk} 
\usepackage{lineno}
\usepackage{tikz}
\usetikzlibrary{spy}

\usepackage{amsmath,amssymb}
\usepackage{hyperref}
\usepackage{cleveref}
\usepackage{booktabs}
\usepackage{subcaption}
\usepackage{bm}

\newcommand{\intd}{\,\mathrm{d}}

\newcommand{\abs}[1]{\left\vert#1\right\vert}
\newcommand{\norm}[1]{\left\Vert#1\right\Vert}
\newcommand{\dotproduct}[2]{\left\langle#1\middle\vert#2\right\rangle}

\let\originalleft\left
\let\originalright\right
\renewcommand{\left}{\mathopen{}\mathclose\bgroup\originalleft}
\renewcommand{\right}{\aftergroup\egroup\originalright}

\newcommand{\imagewithzoom}[5]{\begin{tikzpicture}[spy using outlines={lens={scale=3}, size=1.3cm}]

\node[inner sep=0pt] at (0, 0)  {\includegraphics[width=\textwidth, keepaspectratio]{#1}};
        
\spy [red] on (#2,#3) in node at (#4,#5);
\end{tikzpicture}}

\definecolor{best}{rgb}{1, 0.7, 0.7}
\definecolor{second}{rgb}{1, 0.85, 0.7}

\title[TraM-NeRF]{TraM-NeRF: Tracing Mirror and Near-Perfect Specular Reflections through Neural Radiance Fields}

\author[L. Holland, R. Bliersbach, J. Müller, P. Stotko, R. Klein]{Leif Van Holland, Ruben Bliersbach, Jan U. Müller, Patrick Stotko, Reinhard Klein \\
University of Bonn}
\begin{document}

\teaser{
    \includegraphics[width=0.28\textwidth]{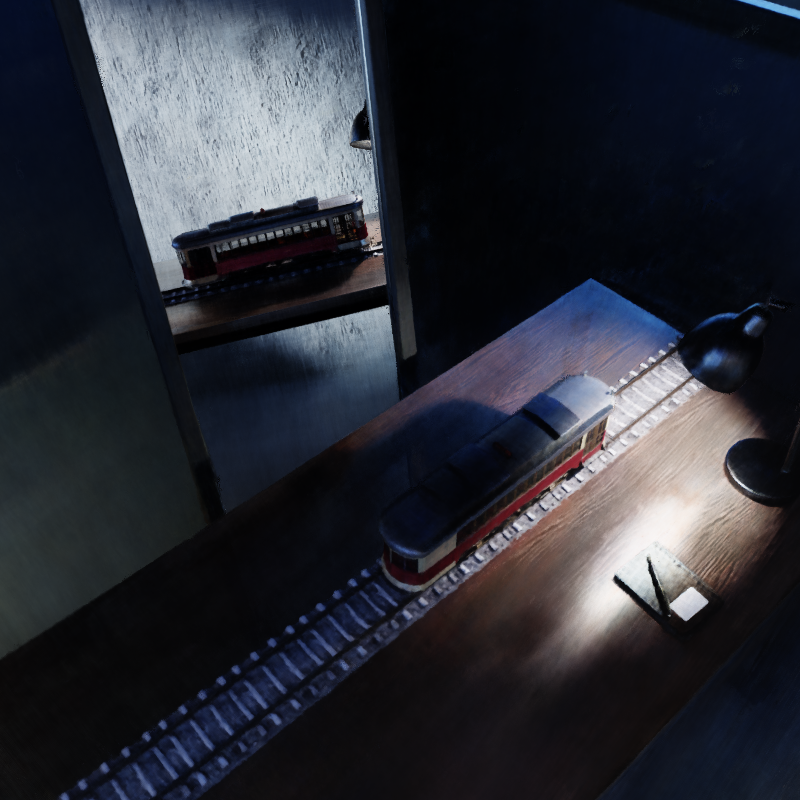}
    \hfill
    \includegraphics[width=0.28\textwidth]{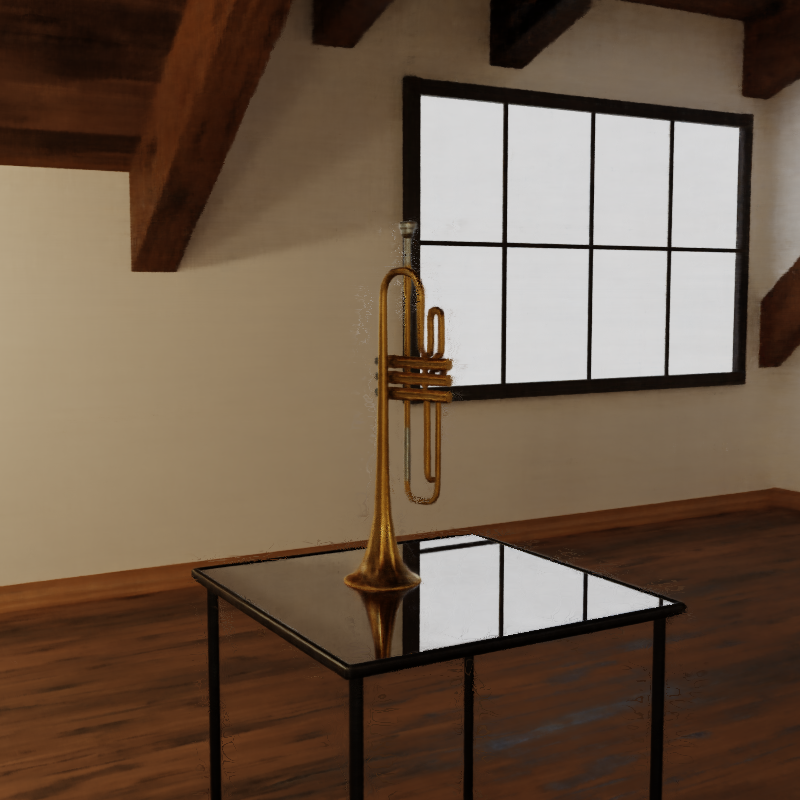}
    \hfill
    \includegraphics[width=0.28\textwidth]{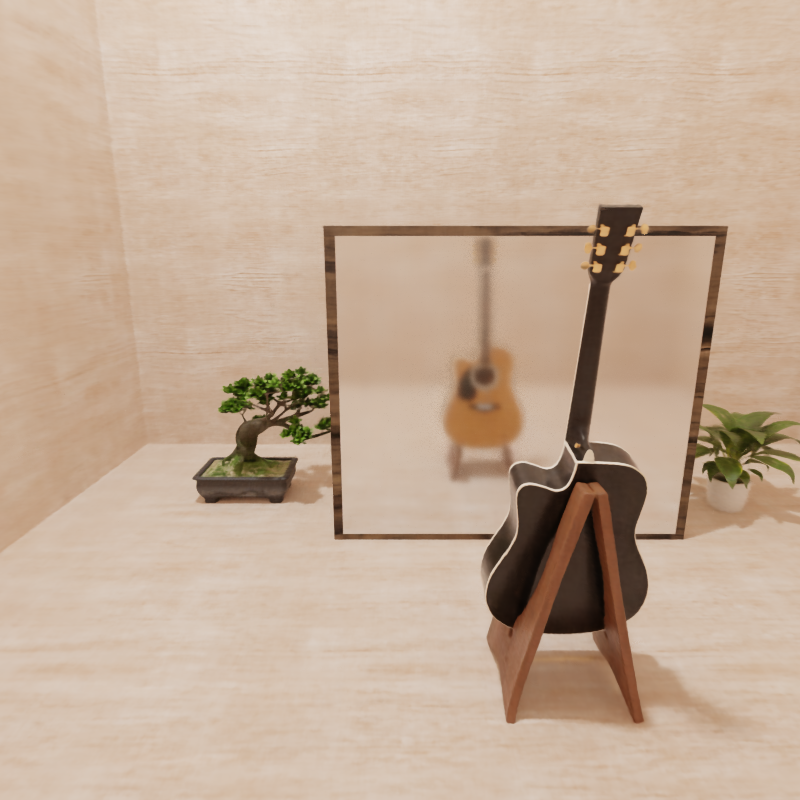}
    \caption{Examples of novel views rendered with our proposed approach on scenes with mirror surfaces (left, center) and near-perfect specular surfaces (right).}
}
\maketitle
\begin{abstract}
    Implicit representations like Neural Radiance Fields (NeRF) showed impressive results for photorealistic rendering of complex scenes with fine details.
    However, ideal or near-perfectly specular reflecting objects such as mirrors, which are often encountered in various indoor scenes, impose ambiguities and inconsistencies in the representation of the reconstructed scene leading to severe artifacts in the synthesized renderings.
    In this paper, we present a novel reflection tracing method tailored for the involved volume rendering within NeRF that takes these mirror-like objects into account while avoiding the cost of straightforward but expensive extensions through standard path tracing.
    By explicitly modeling the reflection behavior using physically plausible materials and estimating the reflected radiance with Monte-Carlo methods within the volume rendering formulation, we derive efficient strategies for importance sampling and the transmittance computation along rays from only few samples.
    We show that our novel method enables the training of consistent representations of such challenging scenes and achieves superior results in comparison to previous state-of-the-art approaches.



\end{abstract}  
\section{Introduction}

3D reconstruction and modeling of real-world scenes has been a major research field for decades and plays a crucial role in a diverse range of applications such as video gaming, movies, advertisement, education as well as AR and VR scenarios.
With the recent emergence of neural scene representations and, especially, Neural Radiance Fields (NeRF)~\cite{mildenhall2020nerf}, a compelling degree of photorealism and immersion of the rendered views has been achieved which inspired many further developments~\cite{zhang2020nerf++,reiser2021kilonerf,barron2021mipnerf,mueller2022instant,wang2022nerf,chen2022hallucinated}.
By combining graphics-based volume rendering with an efficient representation of scene density and radiance using neural networks in terms of multilayer perceptrons (MLP), NeRF enables capturing various effects including view-dependent changes of object appearances or volumetric phenomena like clouds.
However, objects with ideal and near-perfect specular reflection behavior which are often encountered in various scenarios and, in particular, many indoor scenes impose a significant challenge to the representation capabilities of radiance fields as they induce a very specific pattern in the light transport.
For the case of a planar mirror, a symmetric version of the visibile scene parts can be observed which appears to be located behind the mirror and gives the illusion of viewing the respective content through a window.
While, a-priori, this ambiguity results into two plausible and consistent interpretations of the structure of the surrounding environment, additional views directly from behind the mirror object allows to resolve the scenario as the representation of the virtual mirrored scene collides with the observations.
As a consequence, severe artifacts will be introduced in the scene representation of NeRF as the underlying volume rendering approach for rendering traces the primary viewing rays and, in turn, implicitly always prefers the inconsistent interpretation.
Several approaches addressed this issue by decomposing the scene into two or more individually consistent radiance fields~\cite{guo2022nerfren,yin2023msnerf} or employ standard path tracing~\cite{zhang2023nemf,mai2023neural} in combination with an extended volumetric field to infer normal directions and specular reflection probabilities~\cite{zeng2023mirror}.
However, this significantly increases the computational burden both in terms of training performance as well as rendering speed and limits their application into other sophisticated and advanced NeRF approaches.

In this paper, we direct our attention towards an efficient formulation of reflection tracing within the volume rendering procedure of NeRF that can be easily adopted in several NeRF variants to enhance their capabilities in handling mirror-like objects.
To this end, we extend the single-ray absorption volume integration by considering the contributions of reflected radiance towards the observed radiance by the camera, effectively moving our model closer to full physically interpretable light transport in the process.
Our proposed method, referred to as TraM-NeRF, explicitly integrates reflected radiance by first annotating near-specular surfaces.
Combining NeRF volume rendering and ray-tracing with physically plausible materials at intersection points introduces an inductive bias into the training of TraM-NeRF that enables it to learn a single coherent scene representation, even when geometry has only been observed in a reflection. 
Our combined radiance estimator allows us to reduce its variance compared to a standard Monte-Carlo approach without additional computational overhead by decreasing the time spend on transmittance computation along rays.

In summary, the key contributions of this work are:
\begin{itemize}
    \item 
    We present TraM-NeRF, an architecture-agnostic extension of NeRF that efficiently represents scenes with mirror-like surfaces, modeling high-frequency reflections in a physically plausible manner within a single coherent scene representation.
    \item 
    We derive a transmittance-aware formulation of the rendering equation to explicitly model reflected radiance at mirror-like surfaces. Additionally, we introduce efficient strategies for importance sampling and transmittance computation, resulting in a reduction in the number of network evaluations compared to Monte-Carlo estimation.
    
    \item We demonstrate the benefits of our formulation in comparison to previous state-of-the-art methods on a variety of challenging scenes some of which include multiple mirror-like surfaces.
\end{itemize}
The code of our implementation is available at \url{https://github.com/Rubikalubi/TraM-NeRF}.

\section{Related Work}

\begin{figure*}[ht]
    \centering
    \includegraphics[width=\textwidth]{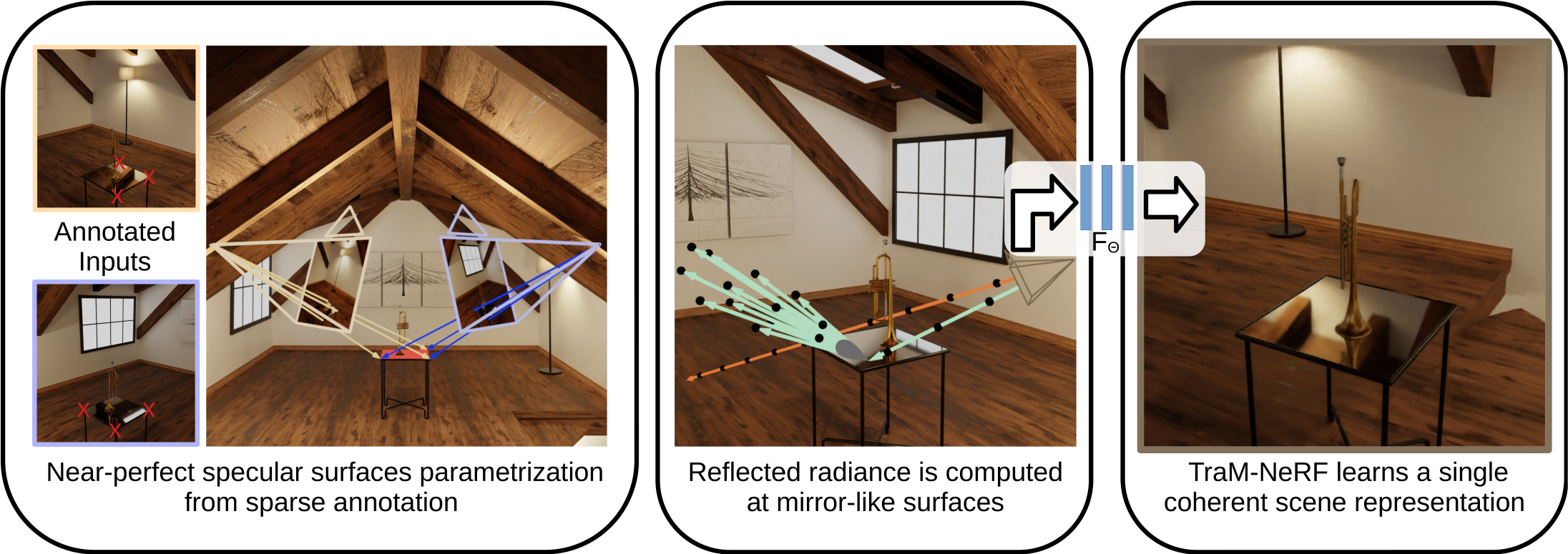}
    \caption{Overview of our proposed method TraM-NeRF. Our approach to parameterize nearly-specular surfaces using only sparse annotations. We introduce a radiance estimator, a crucial component of TraM-NeRF, which combines volume and reflected radiance integration for training and rendering the model. TraM-NeRF learns to represent the observed radiance in a single coherent network.}
    \label{fig:enter-label}
\end{figure*}

\subsection{Neural Scene Representations}

Synthesizing novel views of complex scenes has gained increasing interest due to the promising results achieved with neural scene representations~\cite{lombardi2019neural,sitzmann2019scene,niemeyer2020differentiable,bi2020neural,bi2020deep}.
Among these, especially the work on Neural Radiance Fields~\cite{mildenhall2020nerf} excels in terms of the quality and degree of photorealism of the rendered images and has become very popular, also due to its simple but effective formulation.
In particular, NeRF leverages volume rendering to accumulate the scattered lighting contributions along the traced viewing rays which are represented using volumetric density and view-dependent radiance and parametrized using MLPs.
Various extensions have been developed to further enhance the performance and quality of the original approach such as accelerating the training~\cite{mueller2022instant,chen2022tensorf,fridovich2022plenoxels} as well as the rendering processes~\cite{reiser2021kilonerf,garbin2021fastnerf}, reducing aliasing artifacts by replacing ray-based marching with an integration of 3D conical frustums~\cite{barron2021mipnerf,barron2022mipnerf360}, rendering fine details at very high-resolution~\cite{wang2022nerf,wang20224k,li2023uhdnerf}, or lifting its capabilities to also handle unbounded scenes~\cite{zhang2020nerf++,barron2022mipnerf360} and to reconstruct from in-the-wild image collections~\cite{martin2021nerf,chen2022hallucinated,fridovich2023k} or low dynamic range images with low or varying exposure~\cite{huang2022hdr,mildenhall2022nerf}.

Besides these advances, the underlying representation given by volumetrically baked radiance and density does not account for manipulation tasks like exchanging the environment illumination, so significant effort has been spent into more plausible, physics-inspired scene representations.
Thus, various methods~\cite{zhang2021nerfactor,boss2021nerd,srinivasan2021nerv,boss2022samurai,jin2023tensoir} considered factorizing the radiance field into shape with normals, surface material parameters in terms of a Bidirectional Reflectance Distribution Function (BRDF) as well as environment illumination.
Further approaches~\cite{zhang2021physg,fan2023factored,liang2023envidr,wu2023nefii,ge2023ref} replaced the density-based shape representation by implicit surfaces via signed distance functions (SDF) for a more accurate estimation of the object geometry and normals.

\subsection{Specular Reflections in Neural Representations}

Objects with highly reflective materials often exhibited in captured scenes impose are challenging to reconstruct in decomposed representations of neural radiance fields and have, in turn, attracted increasing attention.
Ref-NeRF~\cite{verbin2022refnerf} reparametrizes the observed radiance based on the local normal vector and its angle to the view direction to a simpler model that shares common structures across multiple views.
PhySG~\cite{zhang2021physg} employs Spherical Gaussians to represent specular reflections in the BRDF which has been later extended by splitting the illumination into a direct and indirect component each modeling an individual specular reflection~\cite{zhang2022modeling}.
Ref-NeuS~\cite{ge2023ref} detect anomalies in the rendered images caused by reflections and incorporate a respective reflection score into the photometric loss as a guidance.
Other works instead directly trace reflections either only in the ideal reflection direction assuming a low material roughness~\cite{liang2023envidr} or using path tracing evaluated via Monte-Carlo estimators~\cite{wu2023nefii}.
Recently, volumetric microflake~\cite{zhang2023nemf} and microfacet~\cite{mai2023neural} fields presented a hybrid rendering approach by combining the ray marching of volume rendering with importance-sampled path tracing according to the distribution of the micro structures.

Most closely related to our work are techniques that explicitly model mirror reflections within the scene.
NeRFReN~\cite{guo2022nerfren} decomposes the scene into two independently traversed and rendered radiance fields consisting of an ordinary NeRF for the transmitted radiance as well as an additional NeRF that only covers the reflected radiance.
The final synthesized image is then obtained by blending the results for the transmitted and reflected fields.
MS-NeRF~\cite{yin2023msnerf} learns radiance and weights into multiple feature fields that are decoded by small MLPs for rendering and then blended together.
Mirror-NeRF~\cite{zeng2023mirror} follows a different direction by representing the scene in a single radiance field and instead further tracing the rays in the ideal reflection direction after hitting a mirror.
The respective normal directions and reflection probabilities used for reflecting the rays are additionally learned in the volumetric neural field similar as in the microflake/microfacet fields~\cite{zhang2023nemf,mai2023neural}.

In contrast to the aforementioned approaches, the primary focus of this work lies in the efficient rendering of unified radiance fields of scenes with mirror and near-perfect specular reflecting objects using only a low number of network evaluations for each importance-sampled reflection ray while achieving a significantly lower variance than standard Monte-Carlo estimators.

\section{Method}
To this end, we start with a brief overview of NeRF. 
Next, we introduce our radiance estimator, a crucial component of TraM-NeRF, which combines volume and reflected radiance integration for rendering and model training. 
We then discuss our approach to parameterize nearly-specular surfaces using sparse annotations.
Finally, we provide implementation and training details for transparency and reproducibility.

\subsection{Neural Radiance Fields}

\begin{figure}
    \centering
    \input{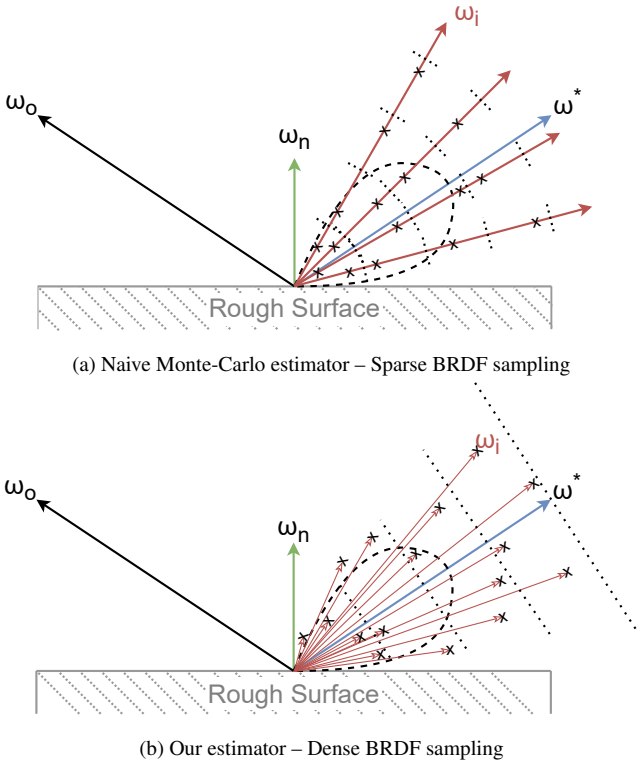}
    \caption{Patterns for BRDF sampling and network evaluation, resulting in different radiance estimators.
    In the standard Monte Carlo approach, the network is evaluate at positions (indicated by cross markers) chosen using stratified sampling for each sampled direction $\omega_i$.
    (b) Our estimator draws directional samples within segments (dashed lines) along the ideal reflection direction $\omega^*$ resulting in a higher angular coverage of the specular lobe with the same number of network evaluations.}
    \label{fig:brdf_sampling}
\end{figure}

We build upon the neural implicit scene representation proposed by \citeauthor*{mildenhall2020nerf} \cite{mildenhall2020nerf} which uses a simple MLP $F_\Theta$ to infer a RGB color value ${c\in\mathbb{R}^3}$ and a density ${\sigma\in\mathbb{R}}$ for a given spatial location ${x\in\mathbb{R}^3}$ and viewing direction ${d\in\mathbb{R}^3}$.
In order to also capture high-frequency details, $x$ is first lifted into a higher-dimensional space using a positional encoding
\begin{equation}
    \gamma(x)
    =
    \left( \sin\big(2^l \pi x\big), \cos\big(2^l \pi x\big) \right)_{l=0}^{L-1}.
\end{equation}

To render an image using $F_\Theta$, we define a camera and consider rays starting at the camera center ${o\in\mathbb{R}^3}$ with direction ${d\in\mathbb{R}^3}$, such that ${ r(t) = o + t \, d }$ represents a point on the ray.
In the original NeRF formulation, the observed radiance 
\begin{equation}
    \label{eq:nerf_render_equation}
    C_e(r)=\int_{t_n}^{t_f} T(t_n \to t) \, \sigma(r(t)) \, c(r(t),d) \intd{t}
\end{equation}
corresponding to ray $r$ is given by the volume integration through an absorbing medium where ${T(t_n \to t)=\exp(-\int_{t_n}^t\sigma(r(s)) \intd{s})}$ represents the accumulated transmittance up to distance $t$, and ${t_n, t_f\in\mathbb{R}}$ describe the near and far plane of $r$.
To be able to compute \cref{eq:nerf_render_equation} in practice, the integral is numerically approximated using quadrature \cite{max1995optical} at locations $t_k$ along the ray, yielding the following discrete sum
\begin{equation}
    C_e(r)\approx \hat{C_e}(r) = \sum_{k=1}^K T_k \,\left(1-e^{-\sigma_k \delta_k}\right) \, c_k.
\end{equation}
Here, $\delta_k = t_k - t_{k-1}$ is the distance between successive locations and ${T(t_n \to t_k) \approx T_k = e^{-\sum_{j=1}^{k-1} \sigma_j \delta_j}}$ approximates the accumulated transmittance.
To prevent using only a discrete subset of locations, we use stratified sampling for the $t_k$, as proposed by \citeauthor*{mildenhall2020nerf} \cite{mildenhall2020nerf}.

During training, the parameters $\Theta$ of the MLP are optimized via gradient descent using a photometric loss $\mathcal{L}$ defined as the mean squared error between the ground-truth colors ${C^*\!(r)}$ and the rendered images $\hat{C}_e(r)$ over a batch of rays $R$:
\begin{equation}
    \mathcal{L} = \frac{1}{\abs{R}} \sum_{r\in R} \norm{C^*\!(r) - \hat{C_e}(r)}_2^2.
\end{equation}

\subsection{Radiance Integration at Near-Perfect Specular Surfaces}
\label{ssec:method_radiance}
Assume that a camera ray $r(t) = o_c + t \, d_c$ intersects with the near-specular surfaces which have been detected in a point $x$.
To allow a model to learn a consistent representation of observed radiance in a single radiance field, we drop the assumption of NeRF that the ray terminates (i.e. the transmittance vanishes) at an opaque surface.
Instead, TraM-NeRF relies on the rendering equation~\cite{kajiya1986rendering} to compute its predicted radiance at intersection points, which states that the radiance $C(x, \omega_o)$ at a point $x$ when observed from direction $\omega_o$ is the sum of the emitted radiance $C_e(x, \omega_o)$ and the reflected radiance $C_r(x, \omega_o)$:
\begin{equation}
   C(x, \omega_o) = C_e(x,\omega_o) + C_r(x, \omega_o).
\end{equation}
Here, the reflected radiance is obtained by evaluating the transport integral 
\begin{equation}
   \label{eq:transport_integral}
   C_r(x, \omega_o) = \int_\Omega f(x, \omega_i, \omega_o) \, C(x, \omega_i) \cos{\theta_i} \intd{\omega_i}
\end{equation}
over the visible hemisphere $\Omega$ where $f(x, \omega_i, \omega_o)$ denotes the BRDF and $\theta_i$ is the angle between the surface normal at $x$ and the direction of incoming light $\omega_i$.
Note that by convention the direction of outgoing light $\omega_o$ faces outwards from $x$.
Consequently, we set the direction of outgoing light to be ${\omega_o = -d_c}$.

\paragraph*{Combining Surface Rendering and Volume Integration}
In order to combine the radiance integration and volume integration, TraM-NeRF assumes that a primary ray scatters into multiple reflected rays at the intersecting point.
Since this ray has passed through an absorbing medium, the combined radiance of the out-branching rays should be attenuated by the transmittance along the intersecting ray.
Whereas NeRF integrates the density along a ray from a starting point close to the camera position up to a point which is chosen a-priori based on the extent of the scene, TraM-NeRF modifies the upper integration bound to stop at the intersection point.
We formalize this concept by introducing a ray length function $\tau(x, \omega)$ which returns the length from the ray origin to the point where it intersects with the detected geometry.
Therefore, we obtain a transmittance-aware version of the rendering equation
\begin{equation}
    C(x, \omega_o) = C_e(x, \omega_o) + T_{\omega_o}(t_n \to \tau(x, \omega_o) ) \cdot C_r(x, \omega_o).
\end{equation}
which takes the attenuation from the absorbing medium into account by multiplying the reflected radiance with the transmittance ${T_{\omega_o}(t_n \to \tau(x, \omega_o))}$.
The emitted radiance observed at point $x$ from direction $\omega_o$
is computed following~\cite{mildenhall2020nerf} by raymarching through the emissive volume until the intersection point is reached:
\begin{equation}
    C_e(x, \omega_o) = \int_{t_n}^{\tau(x, \omega_o)} T_{\omega_o}(t_n \to t) \, \sigma_{\omega_o}(t) \, c_{\omega_o}(t)\intd{t}.
\end{equation}
Note, our modified version retains the offset $t_n$ to prevent double-counting the intersection point in emitted and reflected radiance calculations.
\paragraph*{Monte-Carlo Estimator of Reflected Radiance} 
Our efficient reflected radiance estimator builds upon an established approximation method for the transport integral, employing importance sampling to evaluate a Monte-Carlo estimator.
This estimator is then modified to reduce additional variance introduced when importance sampling a BRDF function, all while keeping the number of network evaluations constant. 
These adjustments enhance the computational efficiency in determining the reflected radiance.

\begin{figure}
    \centering
    \input{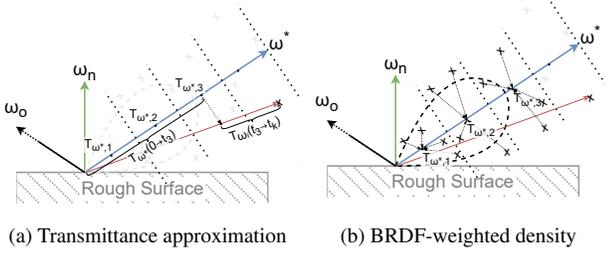}
    \caption{Transmittance approximation used by our estimator. (a) The transmittance ${T_{\omega_i}(t_n \to t_k)}$ in direction $\omega_i$ is approximated using the transmittance ${T_{\omega^*}(t_n \to t_{k-1})}$ along the ideal reflection direction $\omega^*$ for near-perfect specular surfaces. (b) The transmittance in the ideal reflection direction is computed using the BRDF-weighted density of samples within each segment.}
    \label{fig:enter-label}
\end{figure}

Considering the transport integral in \cref{eq:transport_integral}, the respective estimator sampling from a candidate distribution $ p(\cdot) $ is given by
\begin{equation}
    C_r(x, \omega_o) \approx \frac{1}{n} \sum_{i = 1}^n \frac{f(x, \omega_o, \omega_i)}{p(\omega_i)} C(x, \omega_i)
\end{equation}
where we make use of a convenient overlap in notation and re-define $\omega_i$ to be the $i$-th directional sample drawn from $\Omega$ with probability $p(\omega_i)$.
In order to obtain a suitable candidate distribution derived from the BRDF $ f $, we utilize the well-established microfacet theory to model $f$ at the intersection point, assuming roughness arises from a height field of tiny facets distributed according to a distribution $D_\alpha(\cdot)$ with roughness parameter $\alpha$~\cite{cook1982reflectance}.
In particular, we use the widely used GGX reflection model~\cite{walter2007microfacet} which defines the BRDF to be
\begin{equation}
    f(x, \omega_o, \omega_i) = \frac{F(\omega_i, h) \, G(\omega_o, \omega_i, h) \, D_\alpha(h)}{4 \cos{\theta_i} \cos{\theta_o}}
\end{equation}
where $h$ is the half-vector between the incoming and outgoing direction, $\theta_o$ the angle between $\omega_o$ and the surface normal, $F$ is the Fresnel term,
and $G(\cdot)$ a coefficient describing the average attenuation that results from shadowing and masking between microfacets.
For formal definitions of $D_\alpha$, $F$, and $G$, please refer to \cite{walter2007microfacet}.
Utilizing an optics-based analytical GGX reflectance model ensures physical consistency in the learned radiance function and comes with the additional benefit of a well-studied importance sampling technique~\cite{heitz2018sampling, dupuy2023sampling}.
We use visible normal sampling (VNDF) which defines a candidate distribution 
\begin{equation}
    p(h) = \frac{ \max\{0, \dotproduct{\omega_o}{h}\} \, G_1(\omega_o) \, D_\alpha(h)}{4 \cos{\theta_i} \cos{\theta_o}}
\end{equation}
that takes the average attenuation due to mircofacet masking $G_1$ into consideration and has a closed-form sampling routine \cite{heitz2018sampling}.
Note that VNDF is a distribution over the half-vectors $h$ instead of incoming light directions. However, the incoming light direction can be computed via a reflection of the outgoing direction about the half-vector.
Thus, the resulting estimator for the reflected radiance is
\begin{equation}
     C_r(x, \omega_o) \approx \frac{1}{n} \sum_{i=1}^n \underbrace{\frac{F(\omega_i, h) \, G(\omega_o, \omega_i,h)}{G_1(\omega_o)}}_{=\colon f'(\omega_o, \omega_i)} C(x, \omega_i).
\end{equation}

\paragraph*{Efficient Reflected Radiance Approximation}
In order to motivate our efficient radiance approximation, we first review the number of network evaluations required by the estimator when the volume integral is discretized using the same assumption made in NeRF~\cite{mildenhall2020nerf}, that is assuming that the radiance along the ray direction is piece-wise constant.
To simplify both the argument and notation and without loss of generality, we assume that only the camera ray intersects with any detected planar surface.
This simplification combined with the discretized volume integration yields the following reflected radiance estimator:
\begin{equation}
    \begin{split}
    C_r(x, \omega_o) = \frac{1}{n}\sum_{i=1}^n f'(\omega_o, \omega_i) \sum_{k=1}^K T_{\omega_i}(t_n \to t_k) \left(1-e^{-\sigma_{\omega_i,k} \delta_{\omega_i,k}}\right) c_{\omega_i,k}    
    \end{split}
\end{equation}
Figure \ref{subfig:brdf_sparse} provides a visualization of the network evaluation pattern for each directional sample.
A crucial observation is that computing this equation entails $ K $ network evaluations for each direction sampled using VNDF.
To achieve a radiance estimate with minimal noise, a large number of directional samples are required, necessitating a significant number of costly network evaluations.
In light of the aforementioned challenges, we aim to improve the computational efficiency of this procedure in TraM-NeRF.
Our strategy involves increasing the number of directional samples without the need for additional network evaluations.
This optimization leverages the observation that scenes with diffuse or low frequency surfaces reflections can be adequately handled by the standard NeRF model.
However, it encounters difficulties in representing scenes with surfaces which display high-frequency reflections that vary significantly with the viewing direction.
In TraM-NeRF, we combine this observation with the assumptions of scene boundedness and locally smooth network-predicted density.
These assumptions allows our estimator to focus primarily on estimating reflected radiance for near-specular surfaces.
Consequently, our estimator can assume a narrow spread of light directions in the samples.
By combining these insights, we expect that the transmittance remains nearly constant with respect to the sampled directions.
In particular, we approximate the transmittance along all sampled directions $\omega_i$ with the transmittance along the ideal reflection direction $\omega^*$:
\begin{equation}
    T_{\omega_i}(t_n \to t_k) \approx T_{\omega^*}(t_n \to t_k).
\end{equation}
This way, we can interchange the order of the Monte Carlo integration and the NeRF volume integration:
\begin{equation}
    \label{eq:method_our_estimator}
    \begin{split}
    C_r(x, \omega_o) = \frac{1}{n}\sum_{i=1}^n f'(\omega_o, \omega_i) \sum_{k=1}^K T_{\omega_i}(t_n \to t_k) \left(1-e^{-\sigma_{\omega_i,k} \delta_{\omega_i,k}}\right) c_{\omega_i,k} \\
    \approx \sum_{k=1}^K T_{\omega^*}(t_n \to t_k) \frac{1}{n}\sum_{i=1}^n  f'(\omega_o, \omega_i) \left(1-e^{-\sigma_{\omega_i,k} \delta_{\omega_i,k}}\right) c_{\omega_i,k}
    \end{split}
\end{equation}

Intuitively, our estimator divides the ideal reflection ray into $K$ segments and traces transmittance solely along the ideal reflection direction.
Within each segment, we randomly select positions and evaluate $n$ directional samples, each contributing to the calculation only once with their impact attenuated based on the transmittance along the ideal reflection direction, which is depicted in Figure~\ref{subfig:transmittance_approx}.
These positions for evaluating the directional samples are uniformly chosen from the interval $[\frac{t_{k-1} + t_{k}}{2}, \frac{t_k + t_{k+1}}{2}]$.
A visualization of this sampling strategy and its resulting evaluation points is shown in Figure~\ref{subfig:brdf_dense}. 
For a more in-depth explanation and its adaptation to a hierarchical optimization procedure, please refer to Section~\ref{ssec:method_training}.
Our estimator computes transmittance once per segment, enabling us to increase the number of directional samples without requiring additional network evaluations to accumulate transmittance. 
This trade-off balances transmittance precision against directional sample count, reducing noise in reflected radiance.
To further reduce the number of network evaluations, we use an average of BRDF-weighted density predictions per segment to calculate the transmittance
\begin{equation}
    T_{\omega^*}(t_n \to t_k) \approx e^{-\sum_{j=1}^{k-1} \left[\frac{1}{n} \sum_{i=1}^n f'(\omega_o, \omega_i) \, \sigma_{\omega_i,j} \right] \, \delta_{j}}.
\end{equation}
which is visualized in Figure~\ref{subfig:segment_transmittance}.

\subsection{Mirror Parameterization and Annotation}
\label{ssec:method_geometry}
TraM-NeRF infers the position of near-specular planar surfaces from sparsely annotated data, relying only on annotations of corner points marked in a few images.
In practice, we get get sufficiently accurate annotations using only two annotated input images per scene.

We represent planar surfaces as triplets of triangle vertices ${T=(v_1, v_2, v_3),\: v_i\in\mathbb{R}^3}$ which allows for efficient intersection tests with rays in the rendering step~\cite{moeller1997fast}.
Given the screen space annotations of three corners in at least two images and their camera poses, the annotations correspond to rays through the scene.
In particular, the $j$-th annotation of vertex $v_i$ defines a ray ${r_{ij}(t) = o_j + t \, d_{ij}}$ with camera origin $o_j$ and ray direction $d_{ij}$ with ${\norm{d_{ij}}_2 = 1}$.
The estimated 3D location $\hat{v}_i$ of the vertex is then given as the point minimizing the lengths of the orthogonal projection onto each ray:
\begin{equation}
    \hat{v}_i = \min_v \sum_j \norm{v - (o_j + d_{ij} \, (v-o_j) \, d_{ij})}_2^2.
    \label{eq:vertex_location}
\end{equation}
By limiting TraM-NeRF to planar surfaces, we can additionally exploit the property that all vertices of a mirror lie on the same plane.
To increase the robustness against inaccuracies in the annotations, we compute the normal of that plane using principal component analysis applied on the set of annotated vertices of the planar surface.
\subsection{Implementation and Training Details}
\label{ssec:method_training}
To assess and compare our estimator for reflected radiance, TraM-NeRF leverages the NeRF framework~\cite{mildenhall2020nerf} and is implemented using PyTorch~\cite{paszke2019pytorch}.
We chose to use the standard NeRF implementation to ensure a clear comparison of the improvements resulting from our contributions and to avoid potential confusion in the assessment of enhancements attributable specifically to our estimator in comparison to those resulting from different unrelated improvements.
Nevertheless, our estimator is adaptable to various implementations of the radiance field networks, making it compatible with methods that enhance parametrization~\cite{barron2021mipnerf, barron2022mipnerf360} or architecture~\cite{mueller2022instant,chen2022tensorf}.
For training TraM-NeRF, we use a modified version of the NeRF training protocol \cite{mildenhall2020nerf}, using the Adam optimizer with a learning rate of ${10^{-3}}$ without decay.
The Adam hyperparameters remained at their default values: ${\beta_1 = 0.9}$, ${\beta_2 = 0.999}$, and ${\epsilon= 10^{-7}}$.
The model underwent  ${6 \times 10^4}$ iterations of training, with a batch size of $2^{14}$ pixels per iteration.

We apply the sampling process outlined in Section~\ref{ssec:method_radiance} to align with NeRF's hierarchical volume sampling approach~\cite{mildenhall2020nerf} consisting of a coarse and a fine stage.
In the coarse stage, we employ stratified sampling to select points along each ray for network evaluation which involves dividing the ray into $K_c$ equal segments and uniformly sampling a position from each segment.
In the fine stage, additional samples along each ray are generated using inverse transform sampling based on density predictions from the coarse stage which results in an additional set of $K_f$ samples.
Given a set of samples, denoted as $t_1, t_2, \ldots, t_{K}$, where $K$ corresponds to the number of coarse ($K_c$) or fine samples ($K_c + K_f$), TraM-NeRF establishes non-uniformly spaced intervals based on them.
These intervals are defined as $\left[\frac{t_{k-1}+t_{k}}{2}, \frac{t_k + t_{k+1}}{2}\right]$, ensuring that each sample point $t_k$ serves as the center of a specific section of it.
We now generate directional samples by choosing a uniformly distributing length
\begin{equation}
    t_{k,i} \sim U_{\left[\frac{t_{k-1}+t_{k}}{2}, \frac{t_k + t_{k+1}}{2}\right]}
\end{equation}
for each interval $k$ and direction $\omega_i$.
Subsequently, the radiance field is queried for its density and color predictions at specific points
\begin{equation}
    x_{k,i} = x + \frac{t_{k,i}}{\dotproduct{\omega^*}{\omega_i}} \omega_i
\end{equation}
where the dot product between the sampled direction and the ideal reflection direction ensures that $x_{k,i}$ falls within the interval, as illustrated in Figure \ref{subfig:brdf_sparse}.

\section{Experiments}

\begin{figure*}[ht]
    \def\subfigwidth{0.195}
    \def\zoomxshift{-1.05}
    \def\zoomyshift{-1.05}
    \def\zoomxshiftalt{1.05}
    \def\zoomyshiftalt{-1.05}
    \centering
    \begin{subfigure}[b]{\subfigwidth\textwidth}
        \imagewithzoom{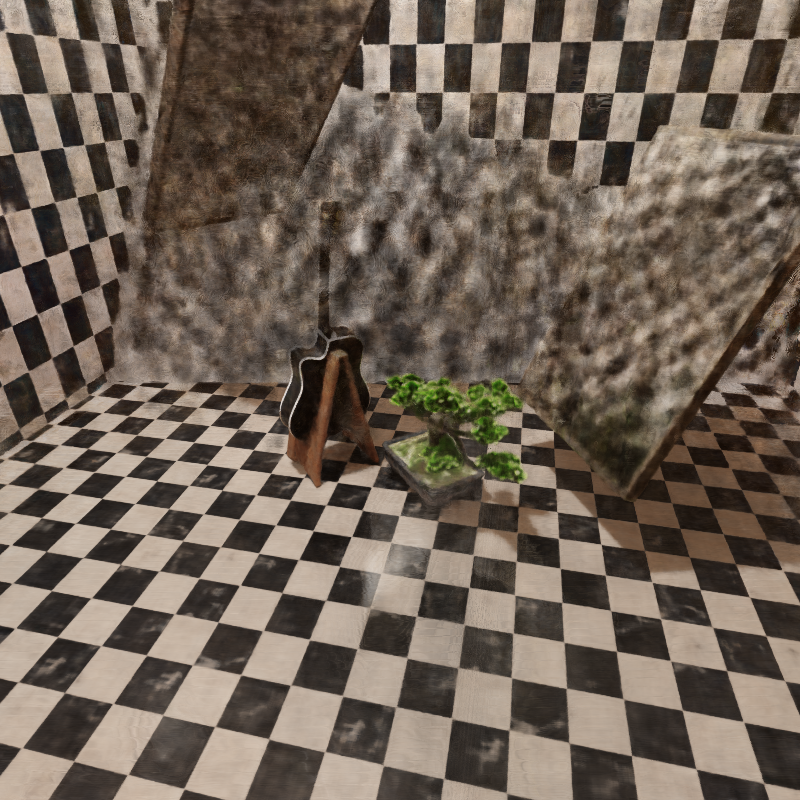}{1.3}{0.7}{\zoomxshift}{\zoomyshift}
        \imagewithzoom{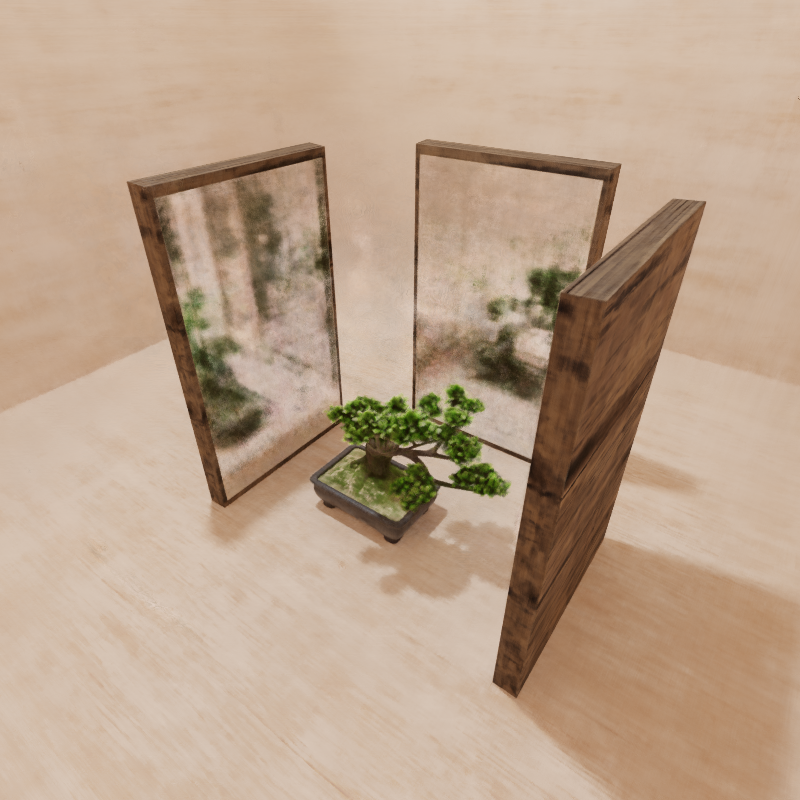}{-0.6}{0.7}{\zoomxshift}{\zoomyshift}
        \imagewithzoom{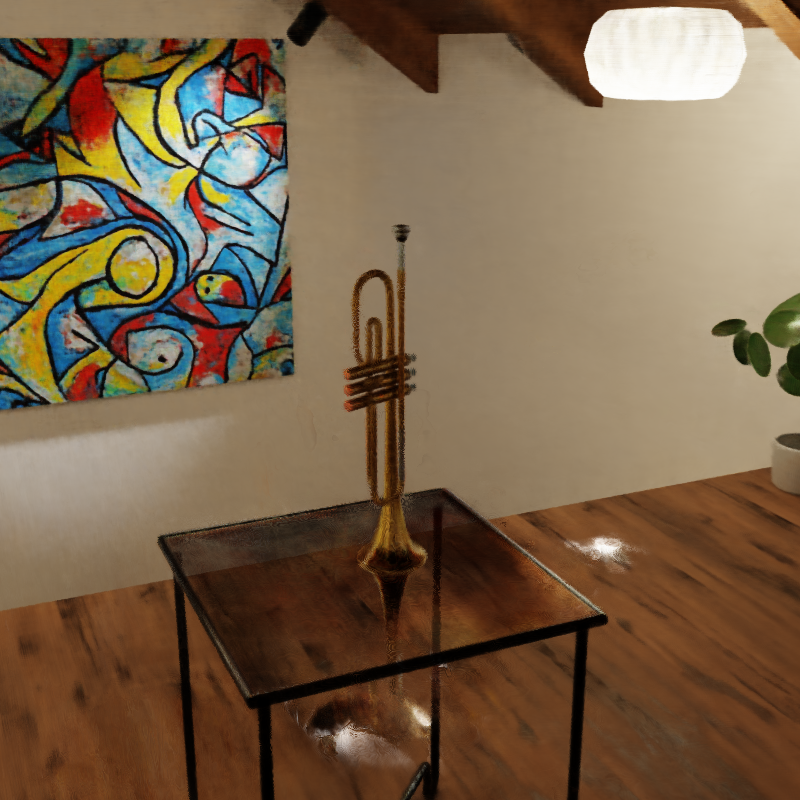}{-0.55}{-0.7}{\zoomxshiftalt}{\zoomyshiftalt}
        \imagewithzoom{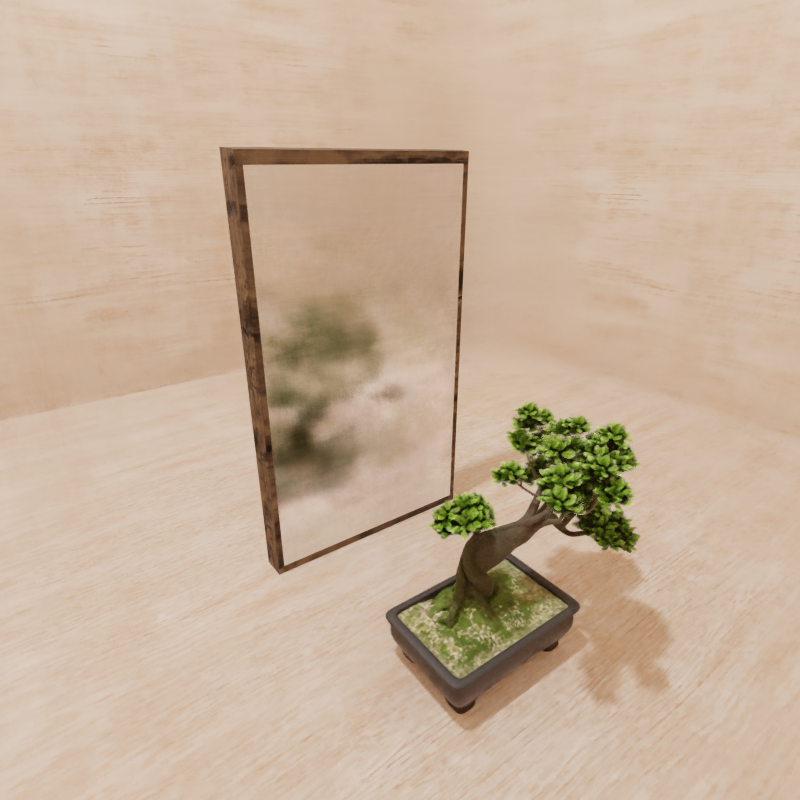}{-0.12}{0.15}{\zoomxshift}{\zoomyshift}
        \caption{Mip-NeRF 360}
    \end{subfigure}
    \hfill
    \begin{subfigure}[b]{\subfigwidth\textwidth}
        \imagewithzoom{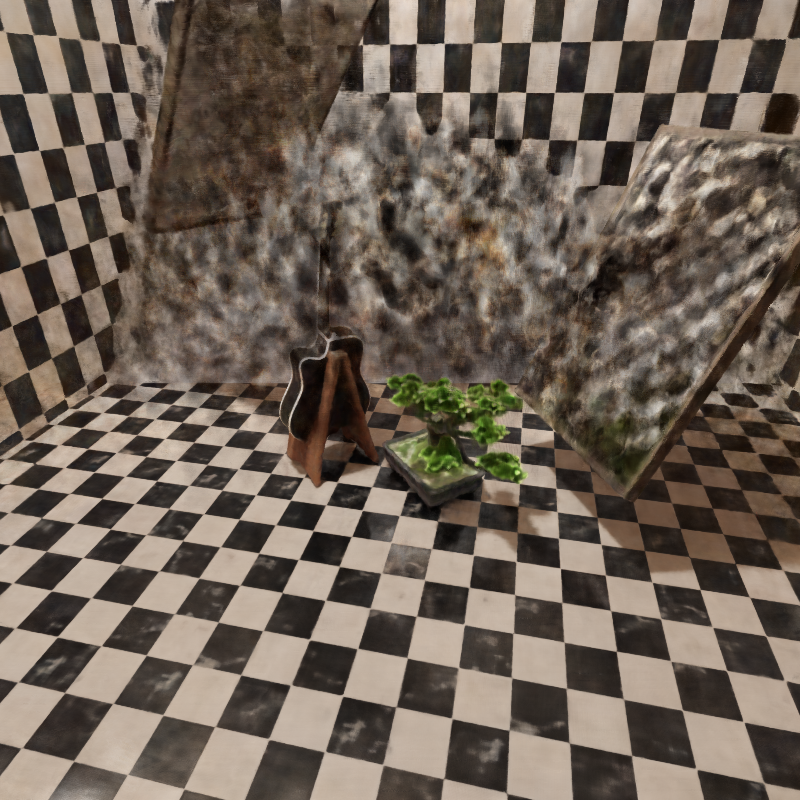}{1.3}{0.7}{\zoomxshift}{\zoomyshift}
        \imagewithzoom{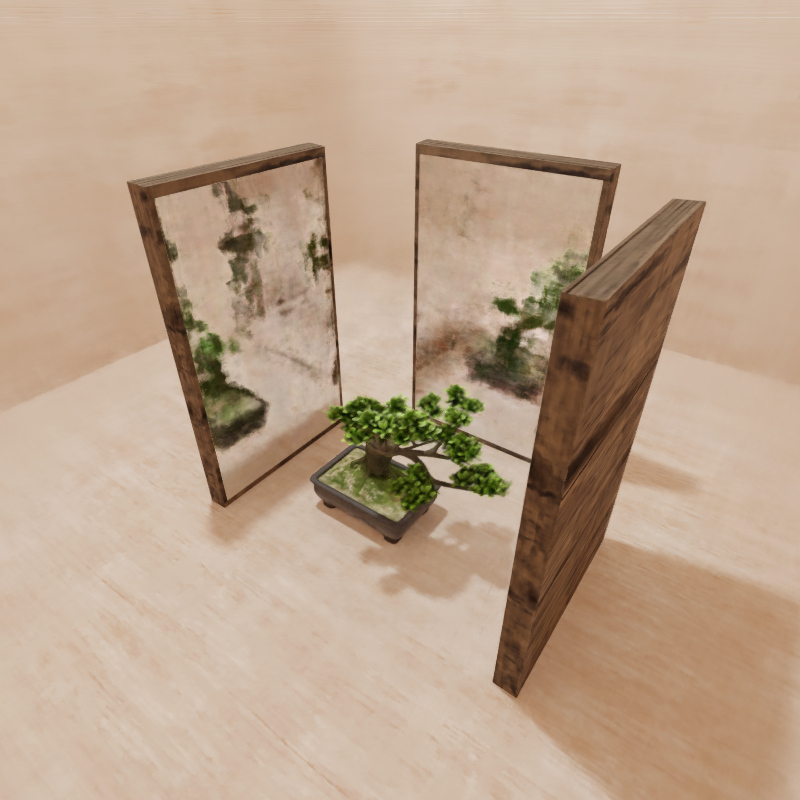}{-0.6}{0.7}{\zoomxshift}{\zoomyshift}
        \imagewithzoom{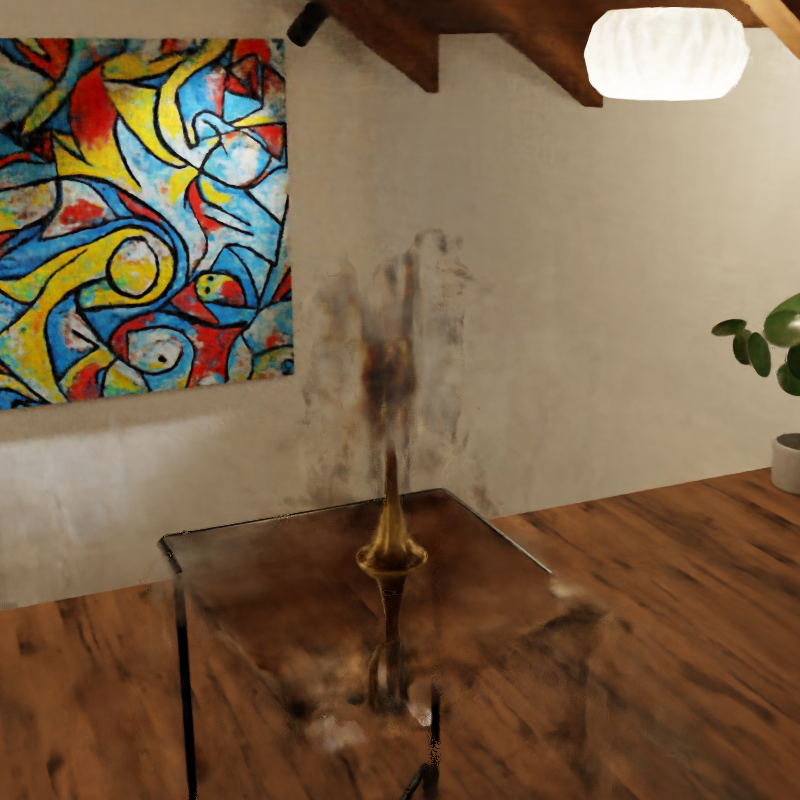}{-0.55}{-0.7}{\zoomxshiftalt}{\zoomyshiftalt}
        \imagewithzoom{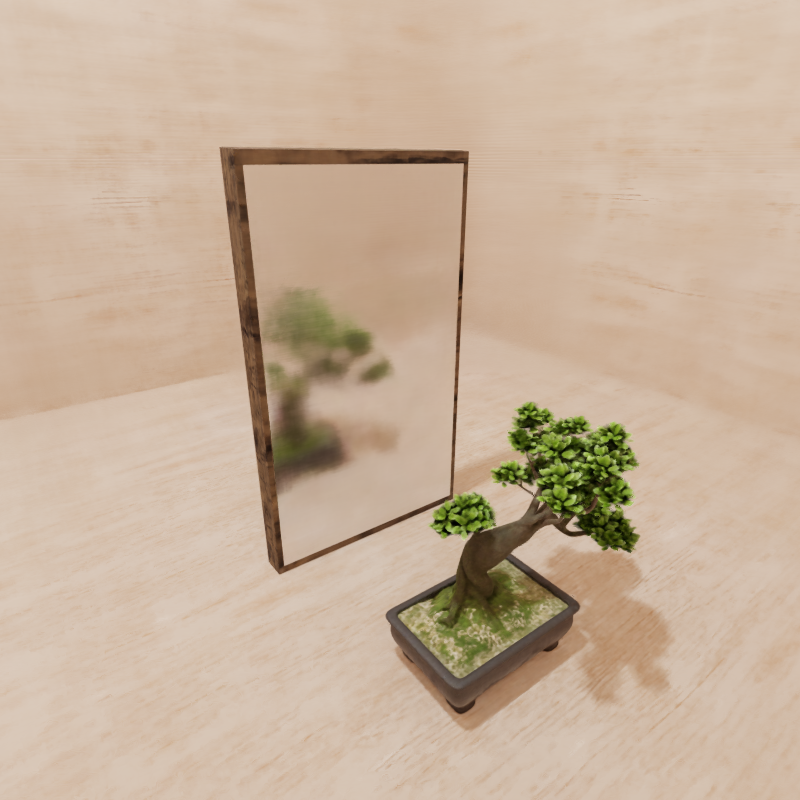}{-0.12}{0.15}{\zoomxshift}{\zoomyshift}
        \caption{Ref-NeRF}
    \end{subfigure}
    \hfill
    \begin{subfigure}[b]{\subfigwidth\textwidth}
        \imagewithzoom{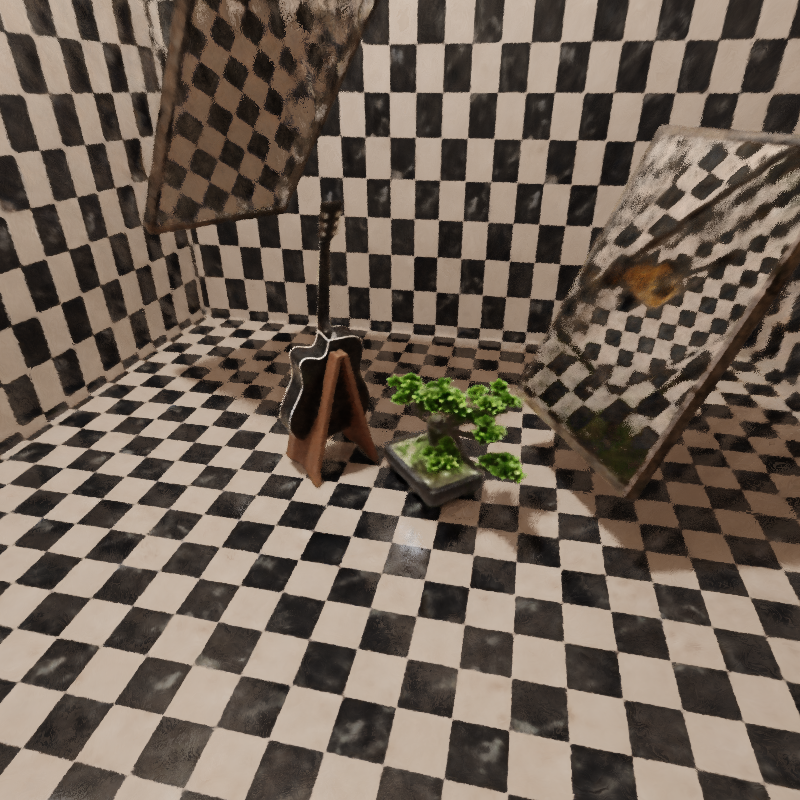}{1.3}{0.7}{\zoomxshift}{\zoomyshift}
        \imagewithzoom{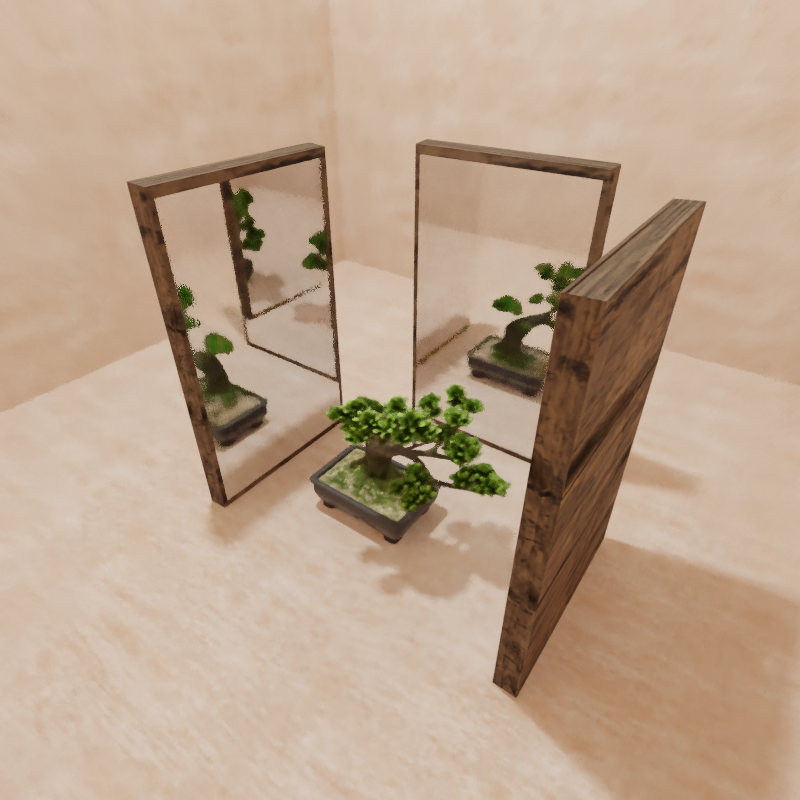}{-0.6}{0.7}{\zoomxshift}{\zoomyshift}
        \imagewithzoom{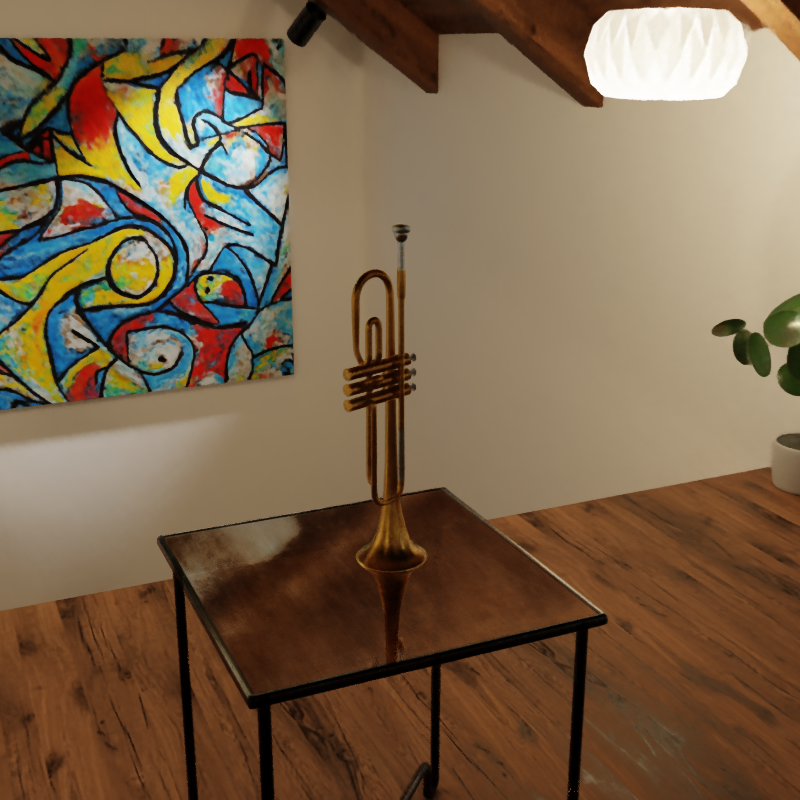}{-0.55}{-0.7}{\zoomxshiftalt}{\zoomyshiftalt}
        \imagewithzoom{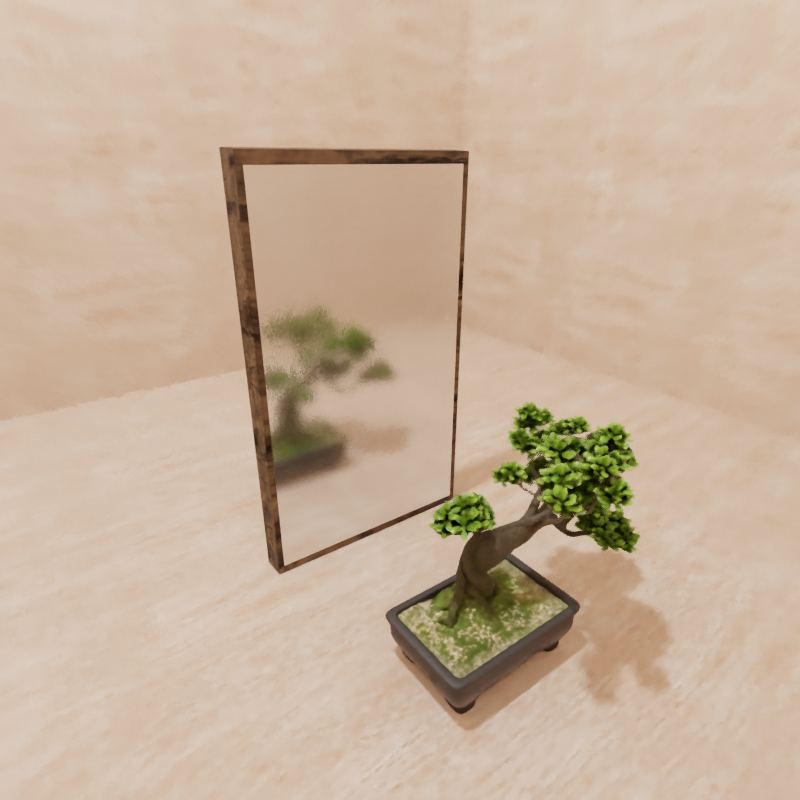}{-0.12}{0.15}{\zoomxshift}{\zoomyshift}
        \caption{MS-NeRF}
    \end{subfigure}
    \hfill
    \begin{subfigure}[b]{\subfigwidth\textwidth}
        \imagewithzoom{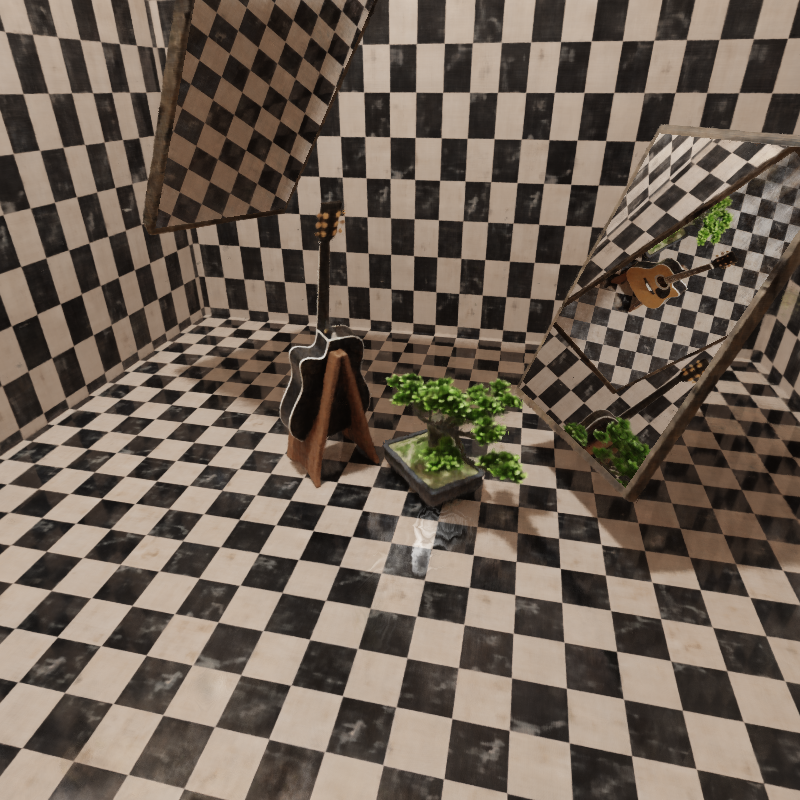}{1.3}{0.7}{\zoomxshift}{\zoomyshift}
        \imagewithzoom{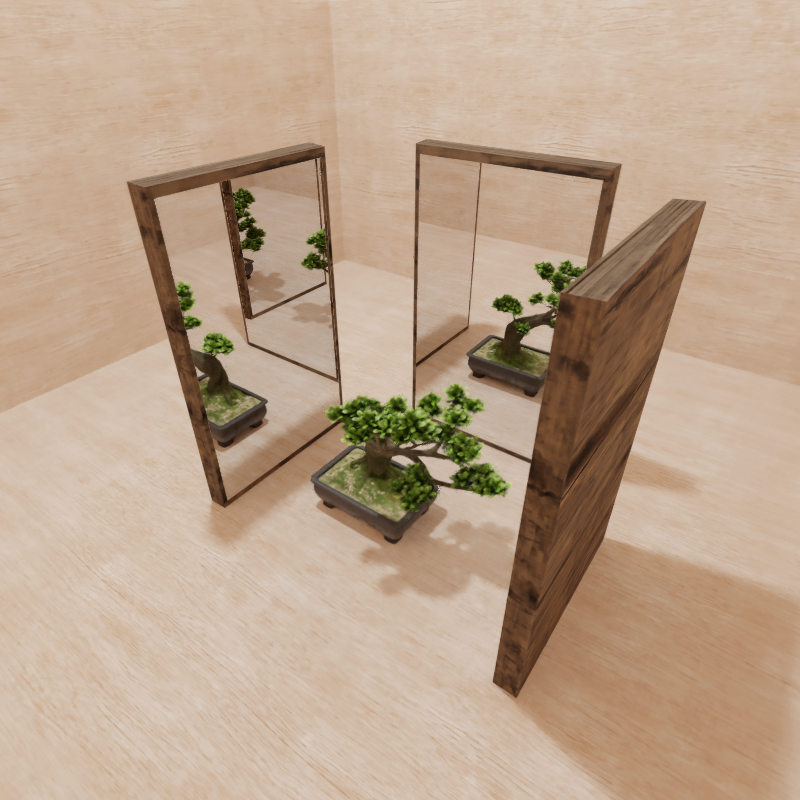}{-0.6}{0.7}{\zoomxshift}{\zoomyshift}
        \imagewithzoom{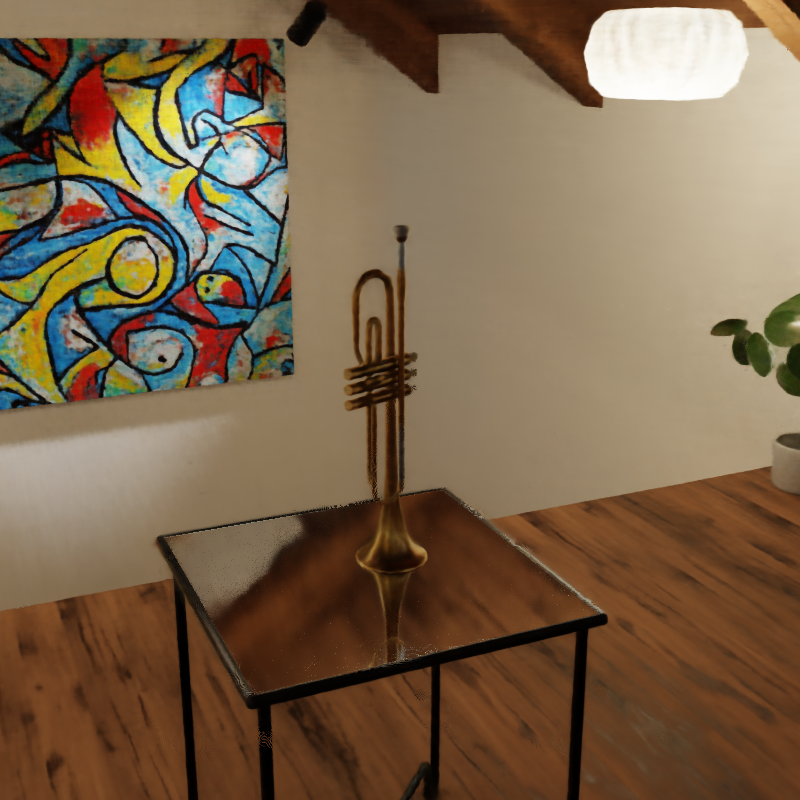}{-0.55}{-0.7}{\zoomxshiftalt}{\zoomyshiftalt}
        \imagewithzoom{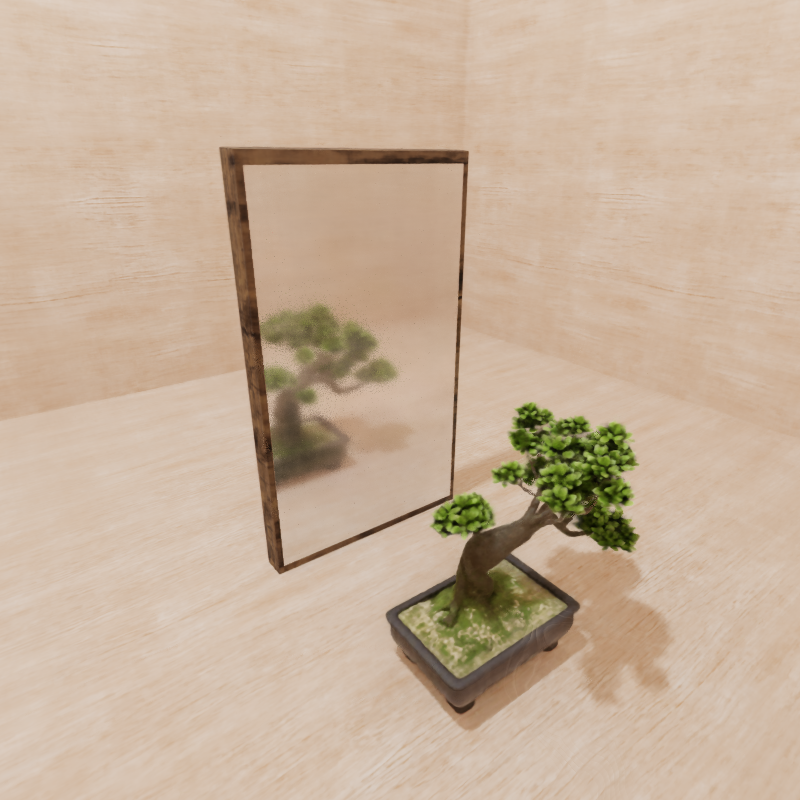}{-0.12}{0.15}{\zoomxshift}{\zoomyshift}
        \caption{Ours}
    \end{subfigure}
    \hfill
    \begin{subfigure}[b]{\subfigwidth\textwidth}
        \imagewithzoom{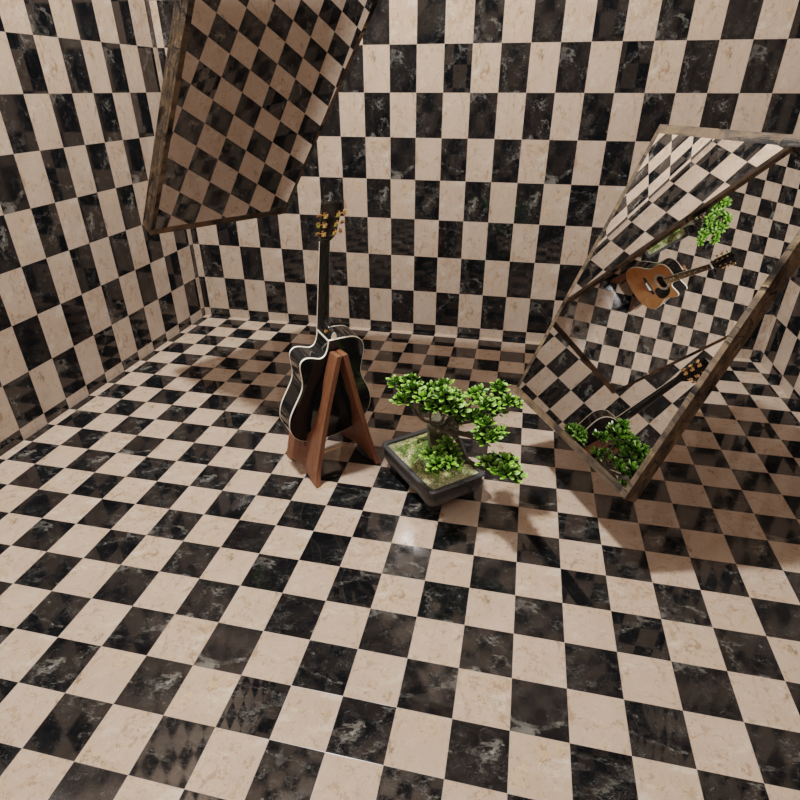}{1.3}{0.7}{\zoomxshift}{\zoomyshift}
        \imagewithzoom{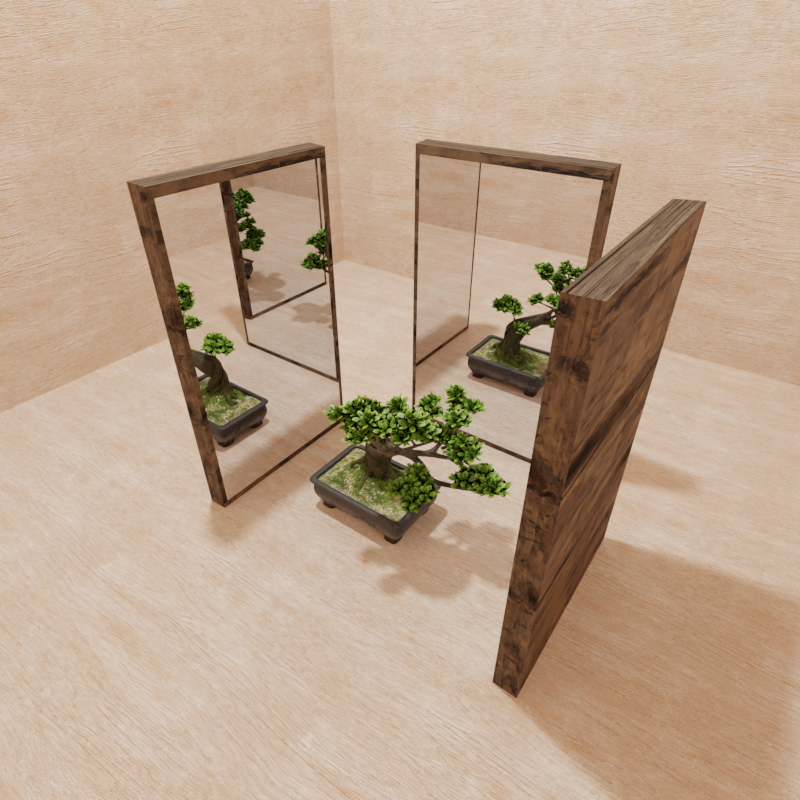}{-0.6}{0.7}{\zoomxshift}{\zoomyshift}
        \imagewithzoom{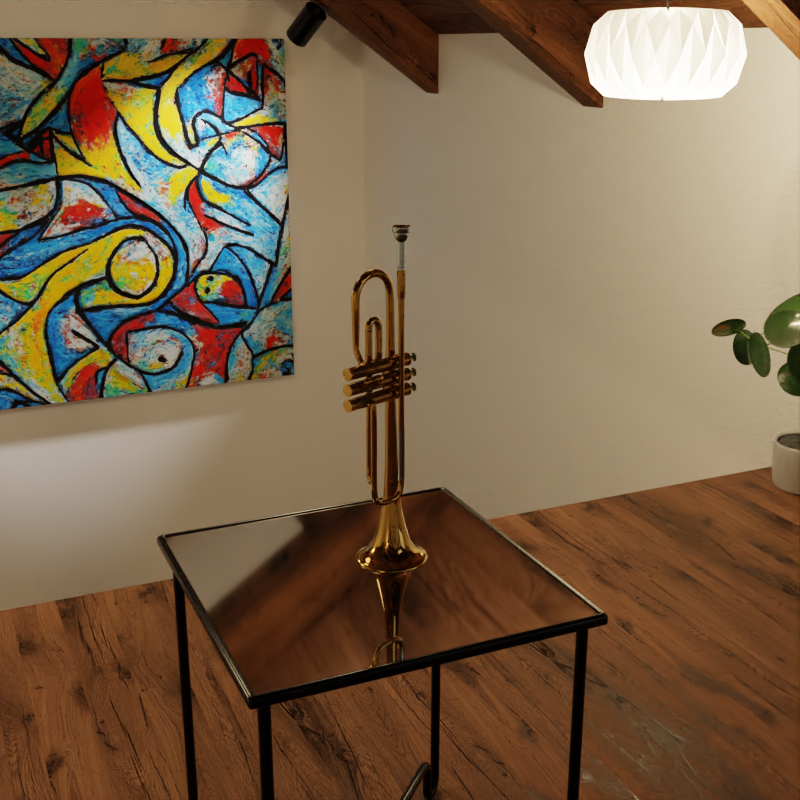}{-0.55}{-0.7}{\zoomxshiftalt}{\zoomyshiftalt}
        \imagewithzoom{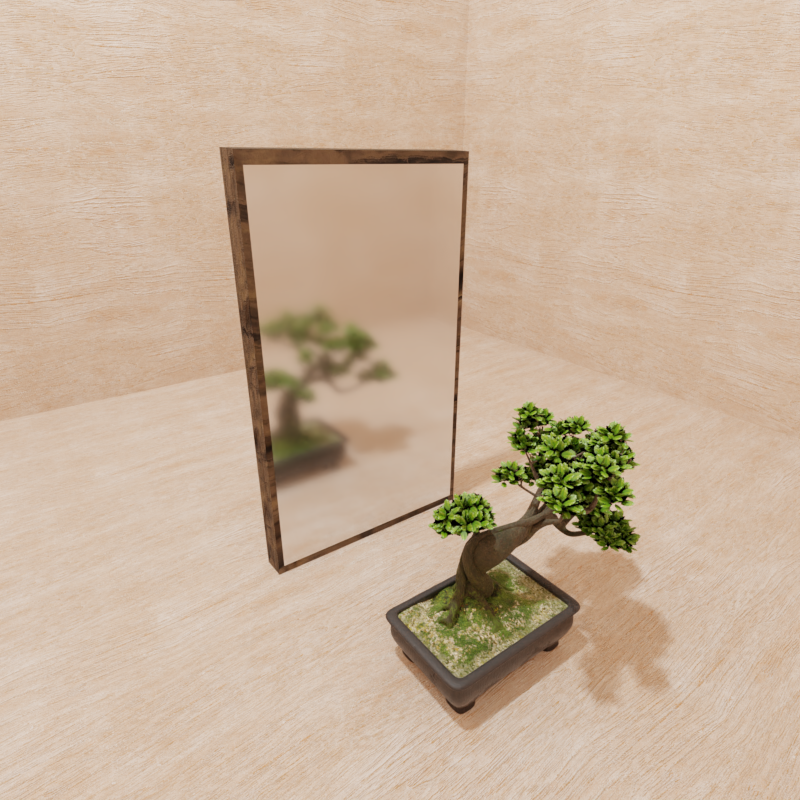}{-0.12}{0.15}{\zoomxshift}{\zoomyshift}
        \caption{Ground Truth}
    \end{subfigure}
    
    \caption{Comparison of different methods on our synthetic dataset with multiple mirrors (rows 1 and 2) and near-perfect specular surfaces (rows 3 and 4). The images shown are views from the test set of the respective scenes. The last column shows the ground truth test image.}
    \label{fig:results}
\end{figure*}

We ran multiple experiments to evaluate different facets of our approach, both on scenes with multiple mirrors, and scenes containing near-specular surfaces.
The scenes are created in the open-source 3D graphics tool Blender to be able to accurately specify the parameters of the materials used.
All 3D models and textures are provided by BlenderKit and CGTrader as royalty free assets.
In total, we created 10 scenes with mirrors and 5 scenes with near-specular surfaces.
For non-forward-facing scenes, we rendered 200 images per scene with cameras sampled from the upper hemisphere around the scene center, looking towards the center.
For evaluation purposes, we withheld 30 per scene from the training process.
As less coverage is required, for the forward-facing scenes we only rendered 100 images per scene and withheld 20 images for evaluation.
We compare our approach on our datasets, both qualitatively and quantitatively, against multiple baseline methods \cite{mildenhall2020nerf,barron2022mipnerf360} and recent methods that explicitly model reflections \cite{verbin2022refnerf,guo2022nerfren,yin2023msnerf}.
To quantify the reconstruction quality, we use the commonly used metrics peak signal-to-noise ratio (PSNR), structural similarity index (SSIM) \cite{wang2004ssim} and learned perceptual image patch similarity (LPIPS) \cite{zhang2018lpips}.

\begin{table*}[]
    \centering
    \fontsize{9pt}{9pt}\selectfont
    \setlength{\tabcolsep}{2pt} 
    \begin{tabular}{lcccccc}
        \toprule
        & \multicolumn{3}{c}{\textbf{Multi-Mirror}} & \multicolumn{3}{c}{\textbf{Near-Specular Surface}} \\
        \cmidrule(lr){2-4} \cmidrule(lr){5-7}
         & PSNR $\uparrow$ & SSIM $\uparrow$ & LPIPS $\downarrow$ & PSNR $\uparrow$ & SSIM $\uparrow$ & LPIPS $\downarrow$ \\
        \midrule
        NeRF \cite{mildenhall2020nerf} & $25.16 \pm 4.70$ & $0.751 \pm 0.112$ & \colorbox{second}{$0.392 \pm 0.091$} & $32.18 \pm 3.01$ & \colorbox{second}{$0.858 \pm 0.006$} & \colorbox{best}{$0.284 \pm 0.045$} \\
        Mip-NeRF 360 \cite{barron2022mipnerf360} & $24.73 \pm 5.98$ & $0.720 \pm 0.147$ & $0.436 \pm 0.087$ & \colorbox{second}{$32.33 \pm 2.83$} & $0.843 \pm 0.012$ & $0.341 \pm 0.015$  \\
        \midrule
        Ref-NeRF \cite{verbin2022refnerf} & $24.37 \pm 5.82$ & $0.726 \pm 0.131$ & $0.444 \pm 0.084$ & $31.75 \pm 4.18$ & $0.825 \pm 0.006$ & $0.378 \pm 0.023$ \\
        MS-NeRF \cite{yin2023msnerf} & \colorbox{second}{$27.61 \pm 6.03$} & \colorbox{second}{$0.767 \pm 0.137$} & $0.405 \pm 0.105$ & $32.29 \pm 2.79$ & $0.831 \pm 0.019$ & $0.400 \pm 0.028$ \\
        \midrule
        Ours & \colorbox{best}{$30.84 \pm 3.54$} & \colorbox{best}{$0.835 \pm 0.053$} & \colorbox{best}{$0.311 \pm 0.086$} & \colorbox{best}{$32.92 \pm 0.85$} & \colorbox{best}{$0.859 \pm 0.019$} & \colorbox{second}{$0.310 \pm 0.013$} \\
        \bottomrule
    \end{tabular}
    \caption{Quantitative comparison of our approach against NeRF baselines and recent works on both multi-mirror scenes and scenes with near-specular surfaces. The metrics are averaged over all test images and across all scenes. \colorbox{best}{Best} and \colorbox{second}{second-best} results are highlighted.}
    \label{tab:results}
\end{table*}

\subsection{Multi-Mirror Scenes}
The first two rows of \Cref{fig:results} show results of various related approaches on scenes with multiple mirrors.
It can be seen that both baseline methods (a) and approaches that consider reflections more explicitly (b, c) struggle to reconstruct higher-order reflections with regard to overall quality (a, b) and high-frequency details (c), while our method can by design represent these regions with the same quality as the rest of the scene.
This strength is also reflected in the qualitative evaluation in \Cref{tab:results}, as our method outperforms the other approaches on all metrics significantly.

\begin{figure}
    \begin{subfigure}[b]{0.235\textwidth}
        \imagewithzoom
        {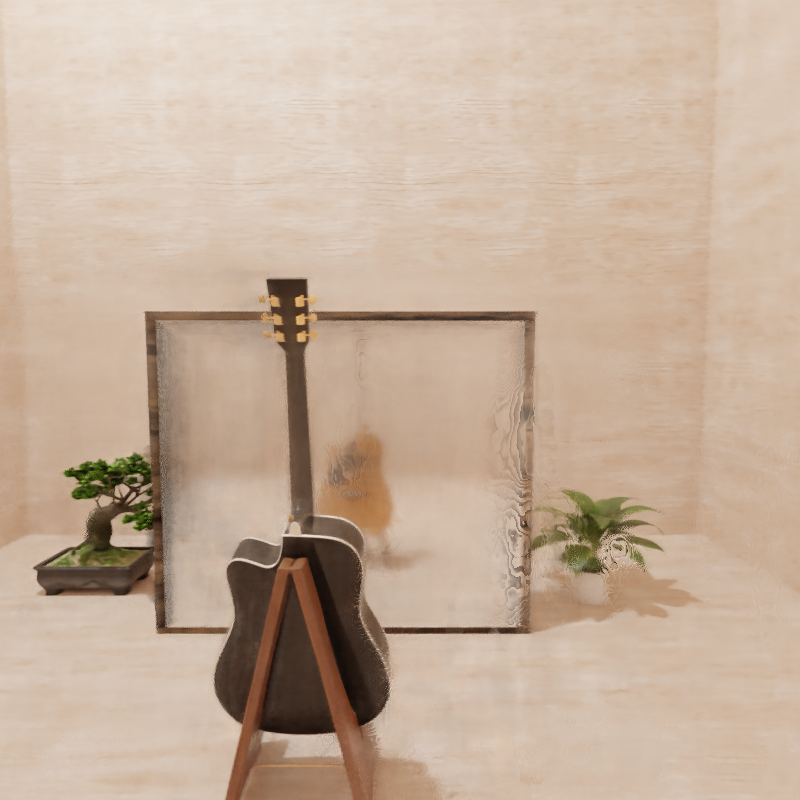}{-0.3}{-0.3}{1.3}{1.3}
        \caption{NeRFReN}
    \end{subfigure}
    \hfill
    \begin{subfigure}[b]{0.235\textwidth}
        \imagewithzoom
        {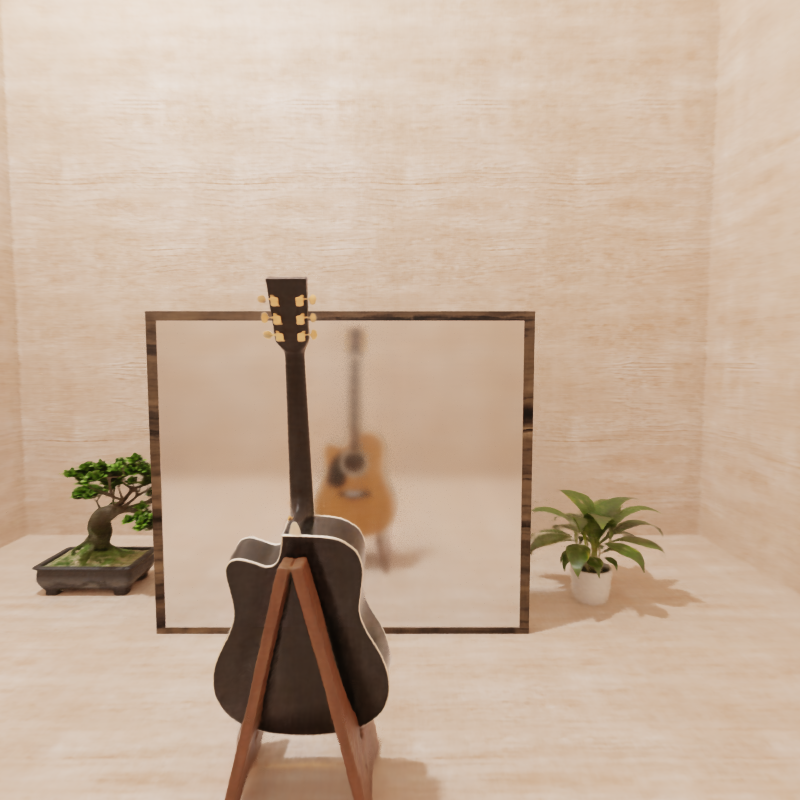}{-0.3}{-0.3}{1.3}{1.3}
        \caption{Ours}
    \end{subfigure}
    \caption{Results on a test view of one of the forward facing scenes that contains a near-specular surface.}
    \label{fig:ff_results}
\end{figure}

\begin{table}[]
    \centering
    \fontsize{9pt}{9pt}\selectfont
    \setlength{\tabcolsep}{2pt} 
    \begin{tabular}{lccc}
        \toprule
         & PSNR $\uparrow$ & SSIM $\uparrow$ & LPIPS $\downarrow$ \\
        \midrule
        NeRF & \colorbox{best}{$36.33 \pm 1.54$} & \colorbox{best}{$0.904 \pm 0.017$} & \colorbox{best}{$0.212 \pm 0.021$} \\
        Mip-NeRF 360 & \colorbox{second}{$35.60 \pm 1.79$} & $0.867 \pm 0.027$ & $0.325 \pm 0.063$ \\
        \midrule
        Ref-NeRF & $35.51 \pm 1.68$ & $0.860 \pm 0.028$ & $0.344 \pm 0.058$ \\
        NeRFReN & $27.38 \pm 8.43$ & $0.778 \pm 0.086$ & $0.489 \pm 0.129$ \\
        MS-NeRF & $34.86 \pm 1.30$ & $0.845 \pm 0.017$ & $0.387 \pm 0.013$ \\
        \midrule
        Ours & $35.42 \pm 1.31$ & \colorbox{second}{$0.885 \pm 0.019$} & \colorbox{second}{$0.271 \pm 0.063$} \\
        \bottomrule
    \end{tabular}
    \caption{Qualitative results on the three forward facing scenes. \colorbox{best}{Best} and \colorbox{second}{second-best} results are highlighted.}
    \label{tab:ff_results}
\end{table}

\subsection{Reflections of Near-Specular Surfaces}

The lower two rows of \Cref{fig:results} show a similar comparison on scenes with near-perfect specular surfaces.
As before, (a) and (b) show a lack of reconstruction quality in regions close to the near-specular surface due to multi-view inconsistencies.
While (c) is able to resolve the inconsistencies, in the third row it can be seen that it fails to learn a clear reflection.
In both the mirror reflections and near-specular surface scenarios, we suppose that the lack of detail in the results of MS-NeRF are due to two effects:
Firstly, \citeauthor*{yin2023msnerf} reduced the sizes of the individual radiance fields to roughly match the size of approaches using a single radiance field.
This leads to less capacity per radiance field in the multi-space formulation.
Secondly, the previous approaches are unable to aggregate information in the reflections from different views consistently, as they are either trying to resolve the multi-view consistencies directly, or move them into a separate radiance field.

\subsection{Forward-Facing Scenes}

In order to compare our results with the approach of \citeauthor*{guo2022nerfren} \cite{guo2022nerfren}, we additionally created three scenes where all camera centers are located on a single plane.
Two scenes contain mirrors, while the surface in the third scene is near-specular.
The default parameters for scenes without manual annotations that are provided by the authors were used for comparison.
We also experimented with providing one or multiple ground-truth masks to their approach, but we found that providing no masks consistently produced the best results.
A qualitative comparison between {NeRFReN} and our approach is shown in \Cref{fig:ff_results}.
It can be seen that our approach is able to better reconstruct details in the near-specular region.
The quantitative comparison additionally shows results of other approaches.
While our approach is not reaching the highest scores on these metrics, it closely follows the first place on the perceptual image metrics and outperforms all other methods that specifically consider reflective surfaces.
This excellent performance of the baseline methods on the forward-facing scenes can be explained by the fact that this scenario does not impose multi-view inconsistencies, which are difficult to resolve using the original NeRF formulation.

\subsection{Reconstruction of Indirectly Observed Regions}

\begin{figure}
    \centering
    \begin{subfigure}[t]{\linewidth}
        \includegraphics[width=\linewidth]{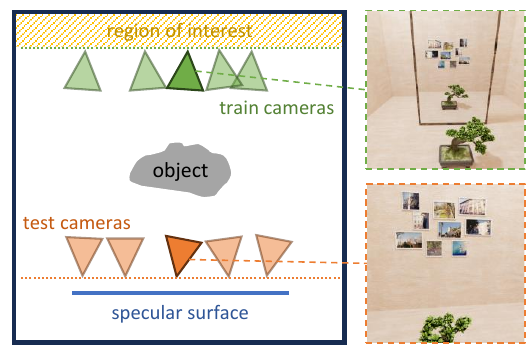}
        \caption{Scene Setup}
    \end{subfigure}
    \par\bigskip
    \begin{subfigure}[t]{0.49\linewidth}
        \includegraphics[width=\linewidth]{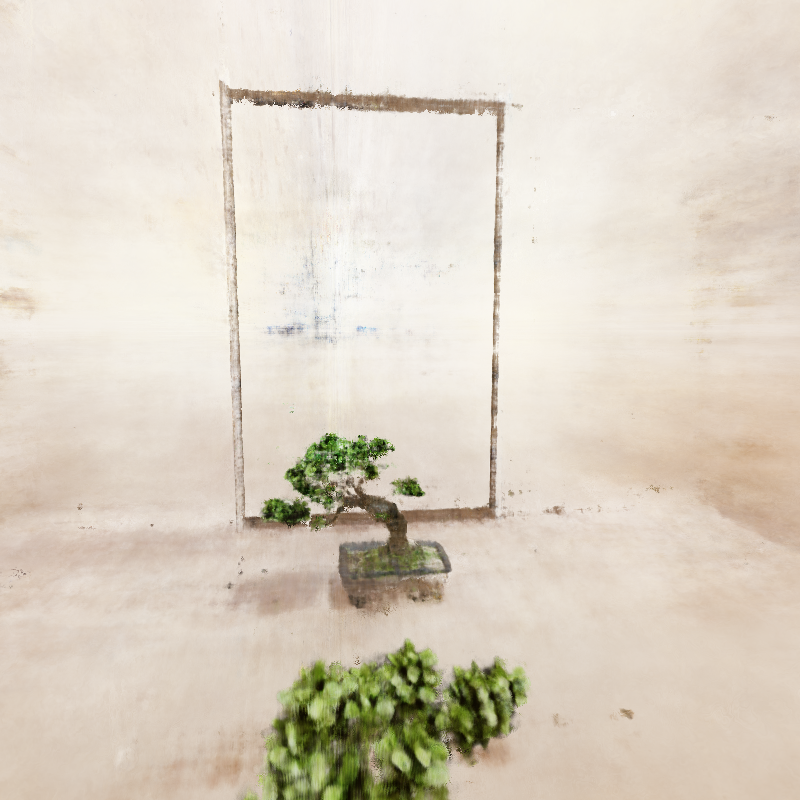}
        \caption{Mip-NeRF 360}
    \end{subfigure}
    \hfill
    \begin{subfigure}[t]{0.49\linewidth}
        \includegraphics[width=\linewidth]{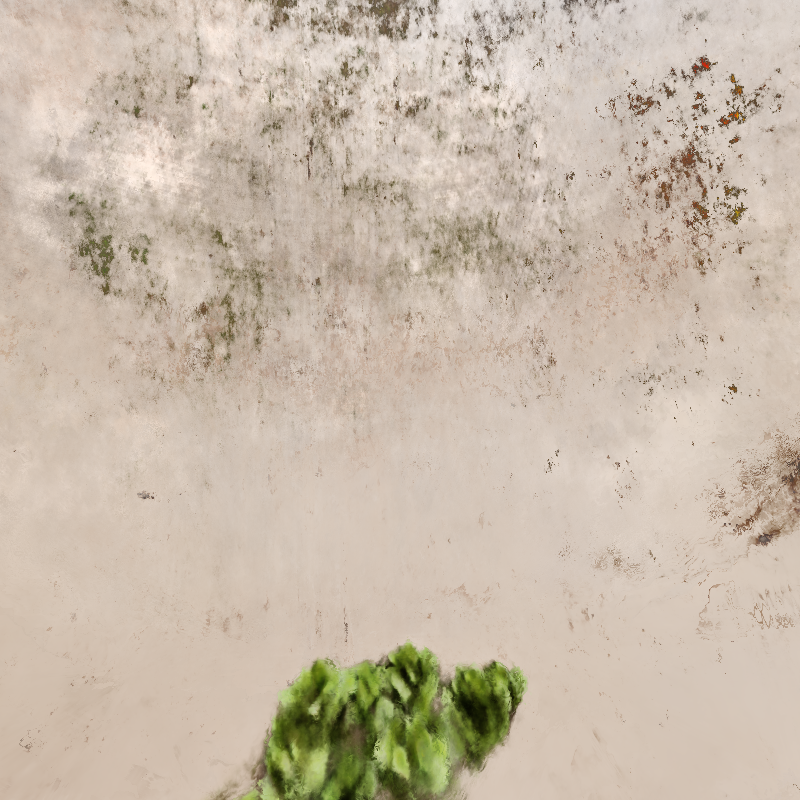}
        \caption{MS-NeRF}
    \end{subfigure}
    \par\bigskip
    \begin{subfigure}[t]{0.49\linewidth}
        \includegraphics[width=\linewidth]{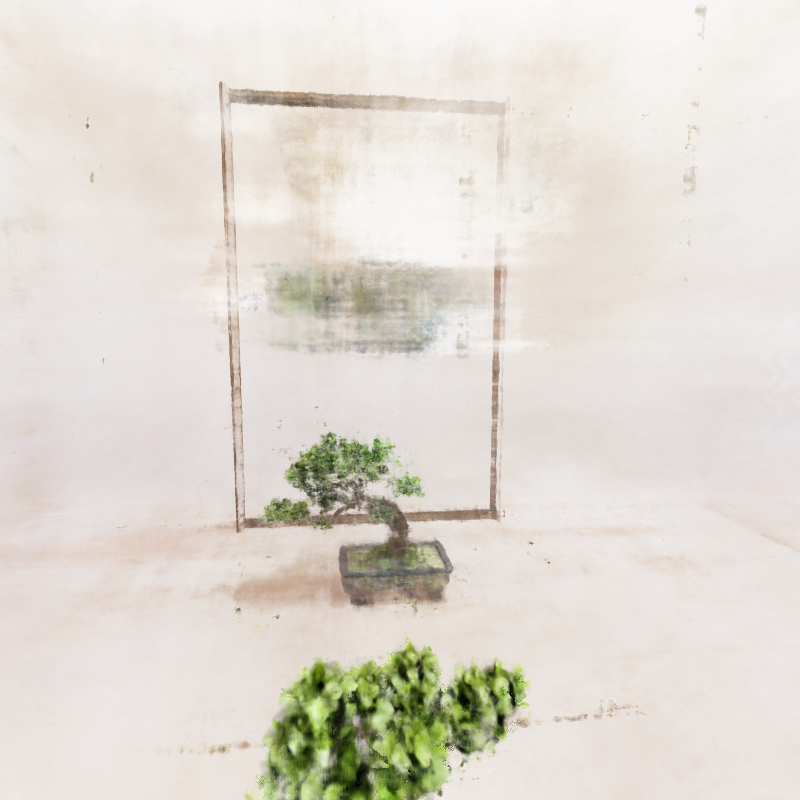}
        \caption{Ref-NeRF}
    \end{subfigure}
    \hfill
    \begin{subfigure}[t]{0.49\linewidth}
        \includegraphics[width=\linewidth]{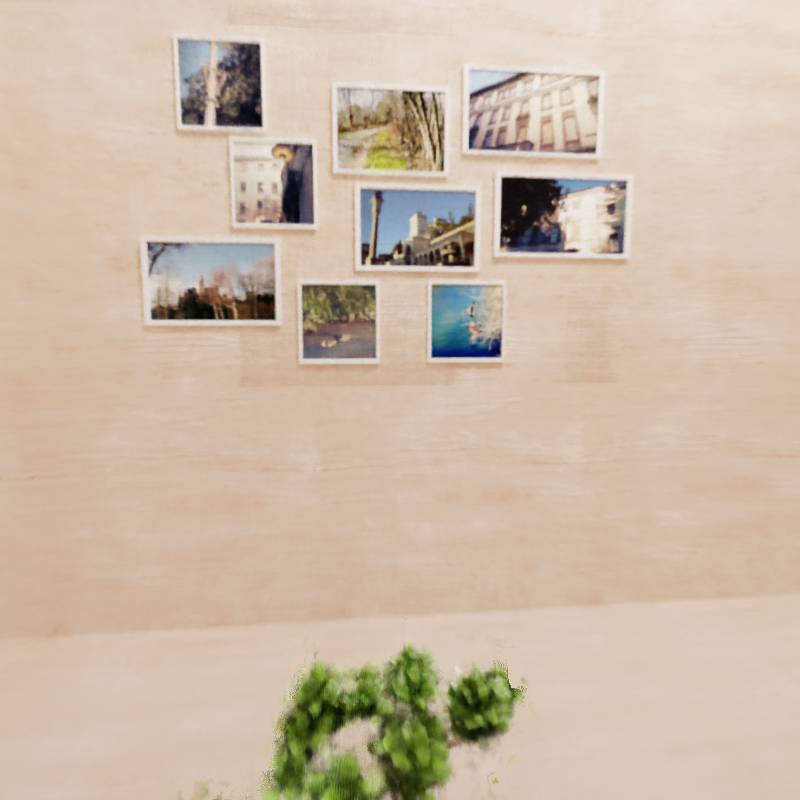}
        \caption{Ours}
    \end{subfigure}
    
    \caption{Experiment with indirectly observed regions. (a) Schematic top view of the scene. Training cameras (green) are placed on a single plane, oriented towards a mirror (blue) that reflects light rays from an unseen region of interest (yellow) towards the training cameras. The test cameras (orange) are placed on a second plane and can directly observe the region of interest. (b)-(e) show the resulting novel views from test cameras produced by previous approaches compared to ours.}
    \label{fig:occlusion_setup}
\end{figure}

One of the advantages of our approach compared to works that model reflections as separate radiance fields \cite{guo2022nerfren,yin2023msnerf} is that information contained in the reflection improves the reconstruction quality of the regions the reflected ray passes through.
To visualize and quantify this we created additional scenes where certain regions of the scene not visible to primary camera rays in any of the training images.
The cameras used to generate the test images are then chosen to cover the regions not seen in the training.
An example of this setup and results are shown in \Cref{fig:occlusion_setup}.
Because the other approaches do not model a change in ray directions, they only extrapolate directly observed scene elements in the unseen regions.
The periodicity in the positional encoding seems to lead to a copy of the observed scene in (b) and (d), while (c) produces noise in the respective regions.
Our approach (e) on the other hand reconstructs high-frequency details that were observable in the reflection of the mirror.

\subsection{Roughness Modification}
\begin{figure}
    \begin{subfigure}[b]{0.15\textwidth}
        \includegraphics[width=\textwidth]{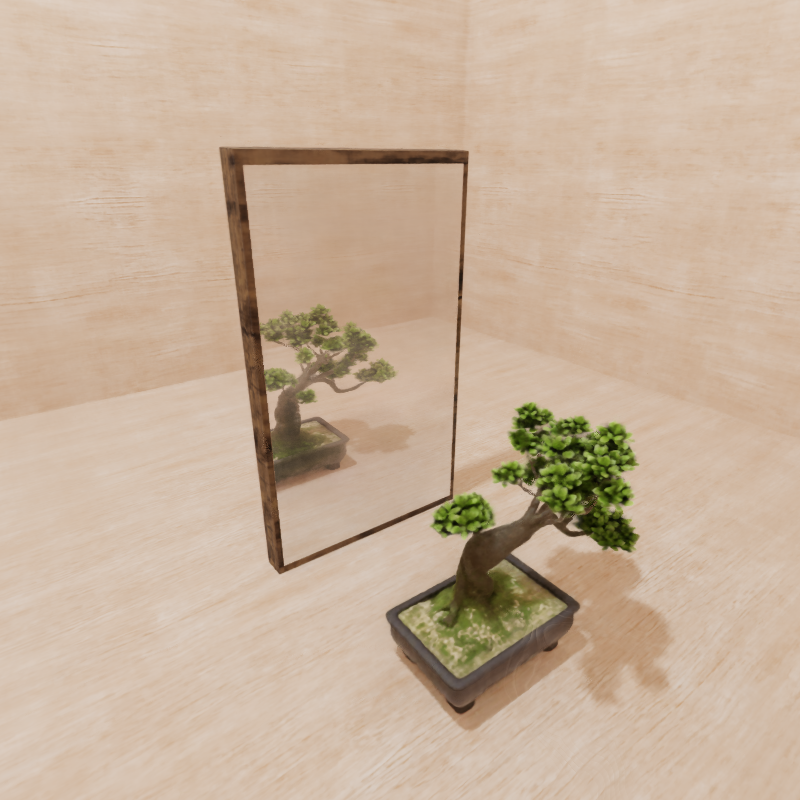}
        \caption{$\alpha = 0$}
    \end{subfigure}
    \hfill
    \begin{subfigure}[b]{0.15\textwidth}
        \includegraphics[width=\textwidth]{images/results/scene_6_ours_r_30.png}
        \caption{$\alpha= 0.09$}
    \end{subfigure}
    \hfill
    \begin{subfigure}[b]{0.15\textwidth}
        \includegraphics[width=\textwidth]{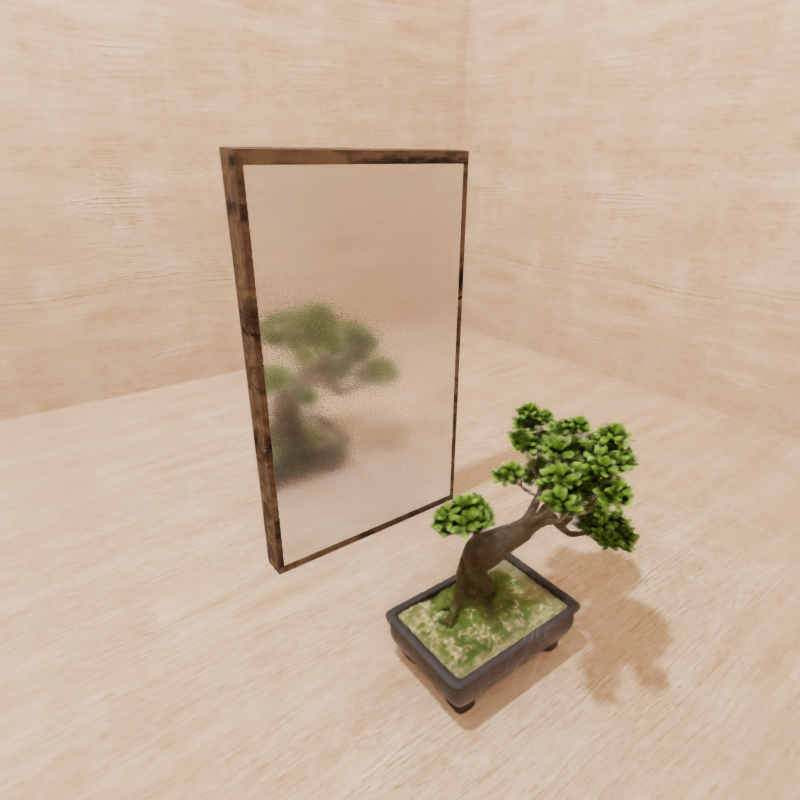}
        \caption{$\alpha=0.018$}
    \end{subfigure}
    \caption{Modification of the roughness parameter $\alpha$ at inference time after being trained on the value shown in (b).}
    \label{fig:roughness_modification}
\end{figure}

In our formulation of volumetric rendering, the BRDF parameters are decoupled from the trained radiance field, which allows us to modify these parameters at inference time.
While we can increase the roughness of a mirror surface after training, we are also able to effectively de-blur a specular surface and produce a mirror-like surface at inference time.
We show this change of roughness at inference time in \Cref{fig:roughness_modification}.

\subsection{Ablation Studies}

We conducted additional experiments to validate some of the design choices of our approach.

\subsubsection{Annotation Robustness}
\label{sec:annotation_robustness}
\begin{figure}
    \centering
    \includegraphics[width=0.9\linewidth]{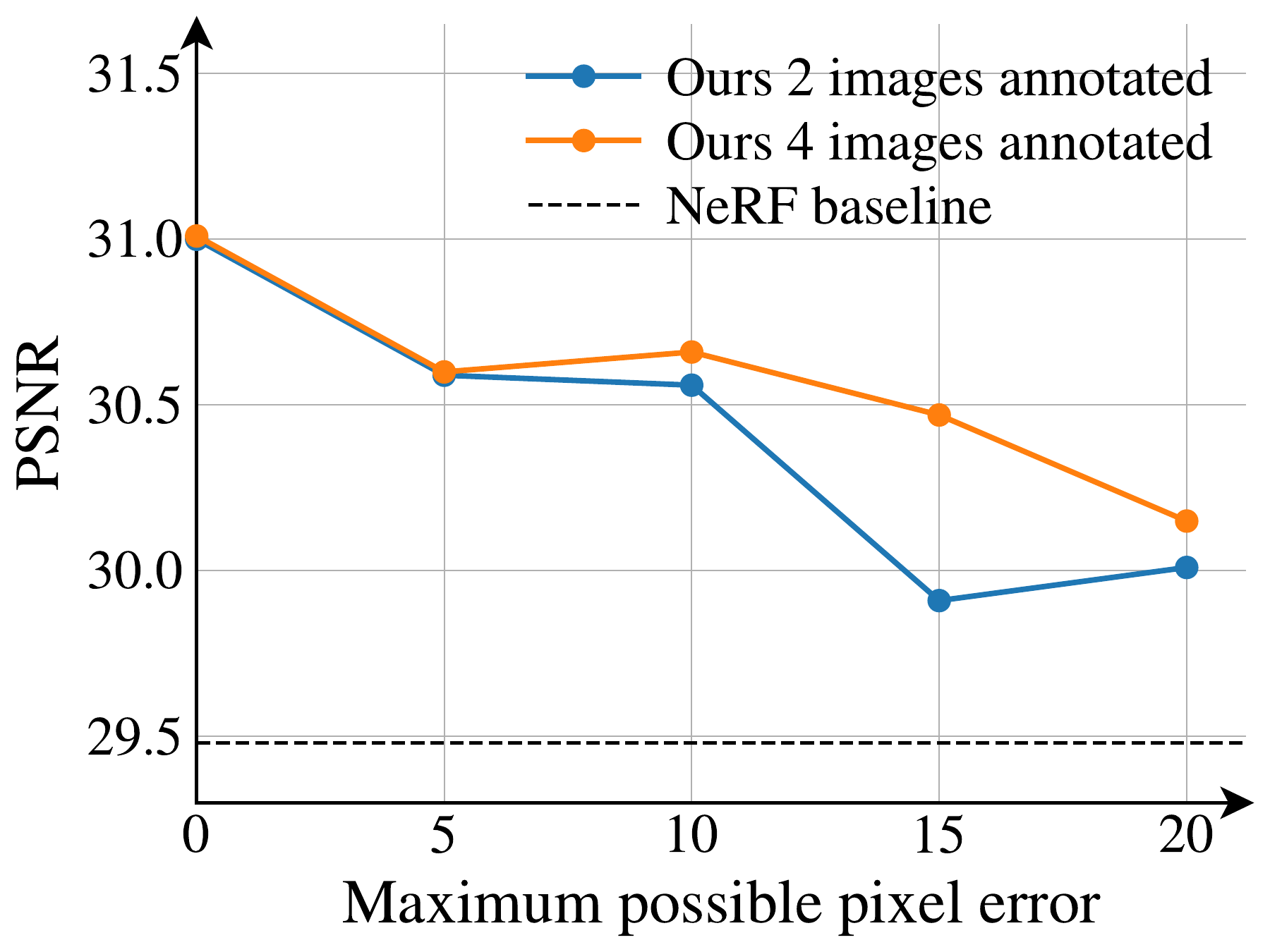}
    \caption{Mean PSNR over all test images of one of the scenes in our dataset after 10000 training iterations, using two (blue) and four (yellow) perfectly annotated images to compute the 3D location of the mirror surface. Different levels of uniform noise are added to simulate different levels of annotation errors. The dashed line shows the resulting mean PSNR of NeRF for the exact annotations after the same amount of training iterations.}
    \label{fig:ablation_robustness}
\end{figure}
To validate our claim that annotations in two images are sufficient to accurately define the position of a rectangular planar mirror in 3D space, we perturbed the annotated positions in screen space with noise sampled from a uniform distribution over the discrete interval $[-k, k]$.
We varied the parameter $k$ to simulate increasingly severe levels of errors in the annotation.
\Cref{fig:ablation_robustness} shows the resulting PSNR of the reconstruction with pixel-exact annotations and different levels of noise on a single scene in our dataset.
It can be seen that carefully annotating only two images yield comparable results to adding two more annotated images.
While higher noise levels in the annotation leads to a drop in reconstruction quality, our approach shows to be robust as it is able to yield an improvement compared to the NeRF baseline with pixel-exact annotations, even under severe noise.

\subsubsection{Modified Ray Sampling}
\begin{figure}
    \centering
    \begin{subfigure}[t]{0.49\linewidth}
        \imagewithzoom{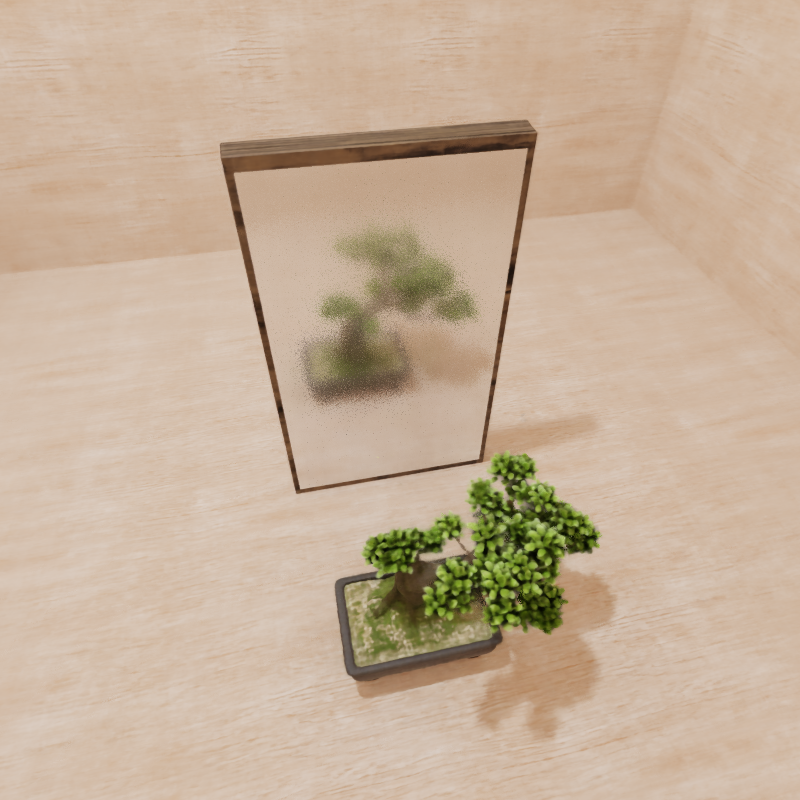}{-0.3}{0.5}{1.3}{1.3}
        \caption{Sparse Sampling}
    \end{subfigure}
    \hfill
    \begin{subfigure}[t]{0.49\linewidth}
        \imagewithzoom{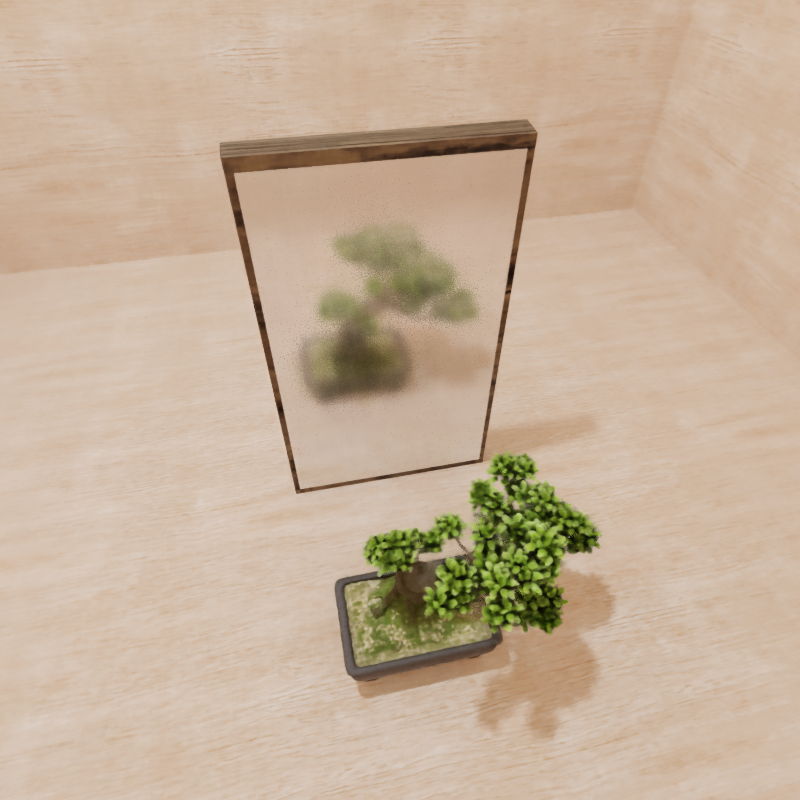}{-0.3}{0.5}{1.3}{1.3}
        \caption{Dense Sampling}
    \end{subfigure}
    \caption{Comparison of sparse and dense sampling variants of our approach on one of the scene of our dataset.}
    \label{fig:ablation_normal_sampling}
\end{figure}
As motivated in \Cref{ssec:method_radiance}, we draw the microfacet normals independently for each sampling location along the main ray (dense) instead of generating multiple rays at the surface intersection and choosing sampling points along these rays (sparse).
We ran an experiment to compare the reconstruction quality of these variants.
\Cref{fig:ablation_normal_sampling} shows that the sparse variant introduces significantly more noise on the near-specular surface than the modified, dense variant.
This, in turn, shows that more samples would be required when using the sparse sampling method to achieve a similar level of noise.

\subsection{Limitations}
While TraM-NeRF achieves promising results for novel view synthesis, it also has some limitations that require further attention. 
In the context of generating novel views, our model inherits NeRF's limitations in extrapolating effectively in areas with insufficient input image coverage, leading to reduced performance.
We observed these hallucinations in parts of the scene that are concealed behind mirrors in a majority of training images.
Additionally, our estimator can overestimate density and transmittance when the assumption of a narrow spread of light directions does not hold.
In this case object reflections can appear larger then expected.
Moreover, the current estimator implementation is limited to single mirror-like surfaces.

\section{Conclusions}
We presented TraM-NeRF, an extension of NeRF that effectively models mirror-like surfaces, accurately capturing high-frequency reflections within a single scene representation. 
By introducing a transmittance-aware variant of the rendering equation for explicit reflection modeling as well as efficient sampling techniques, we are able to reduce he number of network evaluations during ray tracing without increasing the variance.
In the scope of a qualitative and quantitative evaluation, we demonstrated that our techniques outperforms previous methods in challenging scenes with single and multiple mirror-like surfaces.

\section*{Acknowledgements}
This work has been funded by the DFG project KL 1142/11-2 (DFG Research Unit FOR 2535 Anticipating Human Behavior), and additionally by the Federal Ministry of Education and Research of Germany and the state of North-Rhine Westphalia as part of the Lamarr-Institute for Machine Learning and Artificial Intelligence and by the Federal Ministry of Education and Research under grant no. 01IS22094E WEST-AI.


\printbibliography                

@String{CVPR = "IEEE/CVF Conference on Computer Vision and Pattern Recognition (CVPR)"}

@String{ICCV = "IEEE International Conference on Computer Vision (ICCV)"}

@String{ECCV = "European Conference on Computer Vision (ECCV)"}

@String{TVCG = "IEEE Transactions on Visualization and Computer Graphics (TVCG)"}

@String{NeurIPS = "Advances in Neural Information Processing Systems (NeurIPS)"}

@String{TIP = "IEEE Transactions on Image Processing (TIP)"}

@String{TOG = "ACM Transactions on Graphics (TOG)"}

@String{CGF = "Computer Graphics Forum (CGF)"}

@String{EGSR = "Eurographics Symposium on Rendering (EGSR)"}

@String{SIGGRAPH = "Annual Conference on Computer Graphics and Interactive Techniques (SIGGRAPH)"}

@inproceedings{mildenhall2020nerf,
 title={NeRF: Representing Scenes as Neural Radiance Fields for View Synthesis},
 author={Ben Mildenhall and Pratul P. Srinivasan and Matthew Tancik and Jonathan T. Barron and Ravi Ramamoorthi and Ren Ng},
 year={2020},
 booktitle=ECCV,
}

@inproceedings{barron2021mipnerf,
  title={Mip-NeRF: A Multiscale Representation for Anti-Aliasing Neural Radiance Fields},
  author={Barron, Jonathan T and Mildenhall, Ben and Tancik, Matthew and Hedman, Peter and Martin-Brualla, Ricardo and Srinivasan, Pratul P},
  booktitle=ICCV,
  pages={5855--5864},
  year={2021}
}

@article{moeller1997fast,
author = { Tomas   Möller  and  Ben   Trumbore },
title = {Fast, Minimum Storage Ray-Triangle Intersection},
journal = {Journal of Graphics Tools},
volume = {2},
number = {1},
pages = {21-28},
year  = {1997}
}

@inproceedings{barron2022mipnerf360,
  title={Mip-NeRF 360: Unbounded Anti-Aliased Neural Radiance Fields},
  author={Barron, Jonathan T and Mildenhall, Ben and Verbin, Dor and Srinivasan, Pratul P and Hedman, Peter},
  booktitle=CVPR,
  pages={5470--5479},
  year={2022}
}

@inproceedings{verbin2022refnerf,
  title={{Ref-NeRF}: Structured View-Dependent Appearance for Neural Radiance Fields},
  author={Verbin, Dor and Hedman, Peter and Mildenhall, Ben and Zickler, Todd and Barron, Jonathan T and Srinivasan, Pratul P},
  booktitle=CVPR,
  pages={5481--5490},
  year={2022},
  organization={IEEE}
}

@inproceedings{guo2022nerfren,
  title={NeRFReN: Neural Radiance Fields With Reflections},
  author={Guo, Yuan-Chen and Kang, Di and Bao, Linchao and He, Yu and Zhang, Song-Hai},
  booktitle=CVPR,
  pages={18409--18418},
  year={2022}
}

@inproceedings{yin2023msnerf,
  title={Multi-Space Neural Radiance Fields},
  author={Yin, Ze-Xin and Qiu, Jiaxiong and Cheng, Ming-Ming and Ren, Bo},
  booktitle=CVPR,
  pages={12407--12416},
  year={2023}
}

@inproceedings{kajiya1986rendering,
  title={The rendering equation},
  author={Kajiya, James T},
  booktitle=SIGGRAPH,
  pages={143--150},
  year={1986}
}

@article{max1995optical,
  title={Optical models for direct volume rendering},
  author={Max, Nelson},
  journal=TVCG,
  volume={1},
  number={2},
  pages={99--108},
  year={1995},
  publisher={IEEE}
}

@inproceedings{zhang2018lpips,
  title={The unreasonable effectiveness of deep features as a perceptual metric},
  author={Zhang, Richard and Isola, Phillip and Efros, Alexei A and Shechtman, Eli and Wang, Oliver},
  booktitle=CVPR,
  pages={586--595},
  year={2018}
}

@article{wang2004ssim,
  title={Image quality assessment: from error visibility to structural similarity},
  author={Wang, Zhou and Bovik, Alan C and Sheikh, Hamid R and Simoncelli, Eero P},
  journal=TIP,
  volume={13},
  number={4},
  pages={600--612},
  year={2004},
  publisher={IEEE}
}

@article{heitz2018sampling,
  title={Sampling the GGX distribution of visible normals},
  author={Heitz, Eric},
  journal={Journal of Computer Graphics Techniques (JCGT)},
  volume={7},
  number={4},
  pages={1--13},
  year={2018}
}

@inproceedings{walter2007microfacet,
  title={Microfacet models for refraction through rough surfaces},
  author={Walter, Bruce and Marschner, Stephen R and Li, Hongsong and Torrance, Kenneth E},
  booktitle=EGSR,
  pages={195--206},
  year={2007}
}

@inproceedings{zeng2023mirror,
  title={Mirror-NeRF: Learning Neural Radiance Fields for Mirrors with Whitted-Style Ray Tracing},
  author={Zeng, Junyi and Bao, Chong and Chen, Rui and Dong, Zilong and Zhang, Guofeng and Bao, Hujun and Cui, Zhaopeng},
  booktitle={ACM International Conference on Multimedia},
  year={2023}
}

@article{zhang2020nerf++,
	title={Nerf++: Analyzing and improving neural radiance fields},
	author={Zhang, Kai and Riegler, Gernot and Snavely, Noah and Koltun, Vladlen},
	journal={arXiv preprint arXiv:2010.07492},
	year={2020}
}

@article{mueller2022instant,
	author = {Thomas M\"uller and Alex Evans and Christoph Schied and Alexander Keller},
	title = {Instant Neural Graphics Primitives with a Multiresolution Hash Encoding},
	journal = TOG,
	issue_date = {July 2022},
	volume = {41},
	number = {4},
	year = {2022},
	pages = {102:1--102:15},
	articleno = {102},
	numpages = {15},
	publisher = {ACM},
	address = {New York, NY, USA}
}

@inproceedings{chen2022tensorf,
  title={Tensorf: Tensorial radiance fields},
  author={Chen, Anpei and Xu, Zexiang and Geiger, Andreas and Yu, Jingyi and Su, Hao},
  booktitle=ECCV,
  pages={333--350},
  year={2022}
}

@inproceedings{fridovich2022plenoxels,
	title={Plenoxels: Radiance Fields Without Neural Networks},
	author={Fridovich-Keil, Sara and Yu, Alex and Tancik, Matthew and Chen, Qinhong and Recht, Benjamin and Kanazawa, Angjoo},
	booktitle=CVPR,
	pages={5501--5510},
	year={2022}
}

@inproceedings{reiser2021kilonerf,
	title={Kilonerf: Speeding up neural radiance fields with thousands of tiny mlps},
	author={Reiser, Christian and Peng, Songyou and Liao, Yiyi and Geiger, Andreas},
	booktitle=ICCV,
	pages={14335--14345},
	year={2021}
}

@inproceedings{garbin2021fastnerf,
	title={Fastnerf: High-fidelity neural rendering at 200fps},
	author={Garbin, Stephan J and Kowalski, Marek and Johnson, Matthew and Shotton, Jamie and Valentin, Julien},
	booktitle=ICCV,
	pages={14346--14355},
	year={2021}
}

@article{zhang2021nerfactor,
  title={NeRFactor: Neural Factorization of Shape and Reflectance Under an Unknown Illumination},
  author={Zhang, Xiuming and Srinivasan, Pratul P and Deng, Boyang and Debevec, Paul and Freeman, William T. and Barron, Jonathan T.},
  journal=TOG,
  volume={40},
  number={6},
  pages={1--18},
  year={2021},
  publisher={ACM New York, NY, USA}
}

@inproceedings{boss2021nerd,
  title={NeRD: Neural Reflectance Decomposition from Image Collections},
  author={Boss, Mark and Braun, Raphael and Jampani, Varun and Barron, Jonathan T. and Liu, Ce and Lensch, Hendrik},
  booktitle=ICCV,
  pages={12684--12694},
  year={2021}
}

@inproceedings{srinivasan2021nerv,
  title={NeRV: Neural Reflectance and Visibility Fields for Relighting and View Synthesis},
  author={Srinivasan, Pratul P and Deng, Boyang and Zhang, Xiuming and Tancik, Matthew and Mildenhall, Ben and Barron, Jonathan T},
  booktitle=CVPR,
  pages={7495--7504},
  year={2021}
}

@article{boss2022samurai,
  title={Samurai: Shape and material from unconstrained real-world arbitrary image collections},
  author={Boss, Mark and Engelhardt, Andreas and Kar, Abhishek and Li, Yuanzhen and Sun, Deqing and Barron, Jonathan and Lensch, Hendrik and Jampani, Varun},
  journal=NeurIPS,
  volume={35},
  pages={26389--26403},
  year={2022}
}

@inproceedings{jin2023tensoir,
  title={TensoIR: Tensorial Inverse Rendering},
  author={Jin, Haian and Liu, Isabella and Xu, Peijia and Zhang, Xiaoshuai and Han, Songfang and Bi, Sai and Zhou, Xiaowei and Xu, Zexiang and Su, Hao},
  booktitle=CVPR,
  pages={165--174},
  year={2023}
}

@article{fan2023factored,
  title={Factored-NeuS: Reconstructing Surfaces, Illumination, and Materials of Possibly Glossy Objects},
  author={Fan, Yue and Skorokhodov, Ivan and Voynov, Oleg and Ignatyev, Savva and Burnaev, Evgeny and Wonka, Peter and Wang, Yiqun},
  journal={arXiv preprint arXiv:2305.17929},
  year={2023}
}

@inproceedings{liang2023envidr,
  title={ENVIDR: Implicit Differentiable Renderer with Neural Environment Lighting},
  author={Liang, Ruofan and Chen, Huiting and Li, Chunlin and Chen, Fan and Panneer, Selvakumar and Vijaykumar, Nandita},
  booktitle=ICCV,
  year={2023}
}

@article{ge2023ref,
  title={Ref-NeuS: Ambiguity-Reduced Neural Implicit Surface Learning for Multi-View Reconstruction with Reflection},
  author={Ge, Wenhang and Hu, Tao and Zhao, Haoyu and Liu, Shu and Chen, Ying-Cong},
  booktitle=ICCV,
  year={2023}
}

@article{bi2020neural,
  title={Neural Reflectance Fields for Appearance Acquisition},
  author={Bi, Sai and Xu, Zexiang and Srinivasan, Pratul and Mildenhall, Ben and Sunkavalli, Kalyan and Ha{\v{s}}an, Milo{\v{s}} and Hold-Geoffroy, Yannick and Kriegman, David and Ramamoorthi, Ravi},
  journal={arXiv preprint arXiv:2008.03824},
  year={2020}
}

@inproceedings{zhang2023nemf,
  title={NeMF: Inverse Volume Rendering with Neural Microflake Field},
  author={Zhang, Youjia and Xu, Teng and Yu, Junqing and Ye, Yuteng and Jing, Yanqing and Wang, Junle and Yu, Jingyi and Yang, Wei},
  booktitle=ICCV,
  pages={22919--22929},
  year={2023}
}

@inproceedings{mai2023neural,
  title={Neural Microfacet Fields for Inverse Rendering},
  author={Mai, Alexander and Verbin, Dor and Kuester, Falko and Fridovich-Keil, Sara},
  booktitle=ICCV,
  pages={408--418},
  year={2023}
}

@inproceedings{zhang2021physg,
  title={PhySG: Inverse Rendering with Spherical Gaussians for Physics-based Material Editing and Relighting},
  author={Zhang, Kai and Luan, Fujun and Wang, Qianqian and Bala, Kavita and Snavely, Noah},
  booktitle=CVPR,
  pages={5453--5462},
  year={2021}
}

@inproceedings{zhang2022modeling,
  title={Modeling indirect illumination for inverse rendering},
  author={Zhang, Yuanqing and Sun, Jiaming and He, Xingyi and Fu, Huan and Jia, Rongfei and Zhou, Xiaowei},
  booktitle=CVPR,
  pages={18643--18652},
  year={2022}
}

@inproceedings{wu2023nefii,
  title={NeFII: Inverse Rendering for Reflectance Decomposition with Near-Field Indirect Illumination},
  author={Wu, Haoqian and Hu, Zhipeng and Li, Lincheng and Zhang, Yongqiang and Fan, Changjie and Yu, Xin},
  booktitle=CVPR,
  pages={4295--4304},
  year={2023}
}

@inproceedings{martin2021nerf,
  title={Nerf in the wild: Neural radiance fields for unconstrained photo collections},
  author={Martin-Brualla, Ricardo and Radwan, Noha and Sajjadi, Mehdi SM and Barron, Jonathan T and Dosovitskiy, Alexey and Duckworth, Daniel},
  booktitle=CVPR,
  pages={7210--7219},
  year={2021}
}

@inproceedings{chen2022hallucinated,
  title={Hallucinated neural radiance fields in the wild},
  author={Chen, Xingyu and Zhang, Qi and Li, Xiaoyu and Chen, Yue and Feng, Ying and Wang, Xuan and Wang, Jue},
  booktitle=CVPR,
  pages={12943--12952},
  year={2022}
}

@inproceedings{mildenhall2022nerf,
  title={Nerf in the dark: High dynamic range view synthesis from noisy raw images},
  author={Mildenhall, Ben and Hedman, Peter and Martin-Brualla, Ricardo and Srinivasan, Pratul P and Barron, Jonathan T},
  booktitle=CVPR,
  pages={16190--16199},
  year={2022}
}

@inproceedings{fridovich2023k,
  title={K-planes: Explicit radiance fields in space, time, and appearance},
  author={Fridovich-Keil, Sara and Meanti, Giacomo and Warburg, Frederik Rahb{\ae}k and Recht, Benjamin and Kanazawa, Angjoo},
  booktitle=CVPR,
  pages={12479--12488},
  year={2023}
}

@inproceedings{huang2022hdr,
  title={Hdr-nerf: High dynamic range neural radiance fields},
  author={Huang, Xin and Zhang, Qi and Feng, Ying and Li, Hongdong and Wang, Xuan and Wang, Qing},
  booktitle=CVPR,
  pages={18398--18408},
  year={2022}
}

@article{sitzmann2019scene,
	title={Scene representation networks: Continuous 3d-structure-aware neural scene representations},
	author={Sitzmann, Vincent and Zollh{\"o}fer, Michael and Wetzstein, Gordon},
	journal=NeurIPS,
	volume={32},
	year={2019}
}

@inproceedings{niemeyer2020differentiable,
	title={Differentiable volumetric rendering: Learning implicit 3d representations without 3d supervision},
	author={Niemeyer, Michael and Mescheder, Lars and Oechsle, Michael and Geiger, Andreas},
	booktitle=CVPR,
	pages={3504--3515},
	year={2020}
}

@article{lombardi2019neural,
	title={Neural volumes: learning dynamic renderable volumes from images},
	author={Lombardi, Stephen and Simon, Tomas and Saragih, Jason and Schwartz, Gabriel and Lehrmann, Andreas and Sheikh, Yaser},
	journal=TOG,
	volume={38},
	number={4},
	pages={1--14},
	year={2019},
	publisher={ACM New York, NY, USA}
}

@inproceedings{bi2020deep,
	title={Deep reflectance volumes: Relightable reconstructions from multi-view photometric images},
	author={Bi, Sai and Xu, Zexiang and Sunkavalli, Kalyan and Ha{\v{s}}an, Milo{\v{s}} and Hold-Geoffroy, Yannick and Kriegman, David and Ramamoorthi, Ravi},
	booktitle=ECCV,
	pages={294--311},
	year={2020},
	organization={Springer}
}

@inproceedings{wang2022nerf,
  title={NeRF-SR: High Quality Neural Radiance Fields using Supersampling},
  author={Wang, Chen and Wu, Xian and Guo, Yuan-Chen and Zhang, Song-Hai and Tai, Yu-Wing and Hu, Shi-Min},
  booktitle={ACM International Conference on Multimedia},
  pages={6445--6454},
  year={2022}
}

@inproceedings{li2023uhdnerf,
  title={UHDNeRF: Ultra-High-Definition Neural Radiance Fields},
  author={Li, Quewei and Li, Feichao and Guo, Jie and Guo, Yanwen},
  booktitle=ICCV,
  pages={23097--23108},
  year={2023}
}

@article{wang20224k,
  title={4K-NeRF: High fidelity neural radiance fields at ultra high resolutions},
  author={Wang, Zhongshu and Li, Lingzhi and Shen, Zhen and Shen, Li and Bo, Liefeng},
  journal={arXiv preprint arXiv:2212.04701},
  year={2022}
}

@article{dupuy2023sampling,
author = {Dupuy, Jonathan and Benyoub, Anis},
title = {Sampling Visible GGX Normals with Spherical Caps},
journal = CGF,
year={2023}
}

@incollection{paszke2019pytorch,
title = {PyTorch: An Imperative Style, High-Performance Deep Learning Library},
author = {Paszke, Adam and Gross, Sam and Massa, Francisco and Lerer, Adam and Bradbury, James and Chanan, Gregory and Killeen, Trevor and Lin, Zeming and Gimelshein, Natalia and Antiga, Luca and Desmaison, Alban and Kopf, Andreas and Yang, Edward and DeVito, Zachary and Raison, Martin and Tejani, Alykhan and Chilamkurthy, Sasank and Steiner, Benoit and Fang, Lu and Bai, Junjie and Chintala, Soumith},
booktitle = NeurIPS,
pages = {8024--8035},
year = {2019}
}

@article{cook1982reflectance,
  title={A reflectance model for computer graphics},
  author={Cook, Robert L and Torrance, Kenneth E.},
  journal={ACM Transactions on Graphics (ToG)},
  volume={1},
  number={1},
  pages={7--24},
  year={1982},
  publisher={ACM New York, NY, USA}
}

\end{document}